\documentclass{edm_article}
\usepackage{etoolbox}
\usepackage{booktabs} 
\usepackage{graphicx}
\usepackage{wrapfig}
\usepackage{subcaption}
\usepackage{tikz}
\usetikzlibrary{arrows,positioning,shapes}
\tikzset{box/.style={rectangle, draw=black, minimum size=0.25cm}}
\usepackage{amsmath}

\usepackage{hyperref}       
\usepackage{xcolor}
\usepackage{url}            
\usepackage{enumitem}

\newcommand{\figref}[1]{Figure~\ref{#1}}

\usepackage{pifont}
\newcommand{\xmark}{\ding{55}}

\newcommand{\algoRandom}{\ensuremath{\textsc{RandD}}}
\newcommand{\algoEditDist}{\ensuremath{\textsc{EditD}}}
\newcommand{\algoEmbDist}{\ensuremath{\textsc{EditEmbD}}}
\newcommand{\algoTutor}{\ensuremath{\textsc{TutorSS}}}

\newcommand{\algoContinual}{\ensuremath{\textsc{NeurSS}}}
\newcommand{\algoContinualtitle}{N{\footnotesize \textbf{EUR}}SS}
\newcommand{\algoContinualtextbf}{N{\scriptsize \textbf{EUR}}SS}

\newcommand{\algoPCFG}{\ensuremath{\textsc{SymSS}}}
\newcommand{\algoPCFGtitle}{S{\footnotesize \textbf{YM}}SS}
\newcommand{\algoPCFGtextbf}{S{\scriptsize \textbf{YM}}SS}

\newcommand{\benchmark}{\ensuremath{\textsc{StudentSyn}}}
\newcommand{\benchmarktitle}{S{\footnotesize \textbf{TUDENT}}S{\footnotesize \textbf{YN}}}

\usepackage{environ}
\usepackage{esvect} 
\usepackage[ruled,vlined]{algorithm2e}
\NewEnviron{boxcode}[3]{
    \begin{center}
   	 	\scalebox{#2}{
	   	 	\setlength\tabcolsep{4pt}
	   	 	\renewcommand{\arraystretch}{#3}
    		\begin{tabular}{|p{#1}|}    
    			\hline
    			\vspace{0.1mm}    
				\BODY
    			\\\hline
    		\end{tabular}
    	}
    \end{center}
}

\newcommand{\textcode}[1]{{\fontfamily{cmtt}\selectfont #1}\xspace}

\newcommand{\task}{\text{\textcode{T}}}
\newcommand{\code}{\text{\textcode{C}}}

\newcommand{\taskspace}{\ensuremath{\mathbb{T}}}
\newcommand{\codespace}{\ensuremath{\mathbb{C}}}

\newcommand{\DSLMove}{\textcode{move}}
\newcommand{\DSLTurnLeft}{\textcode{turnLeft}}
\newcommand{\DSLTurnRight}{\textcode{turnRight}}

\newcommand{\DSLRepeatUntil}{\textcode{\textsc{RepeatUntil}}}
\newcommand{\DSLIf}{\textcode{\textsc{If}}}
\newcommand{\DSLIfElse}{\textcode{\textsc{IfElse}}}
\newcommand{\DSLElse}{\textcode{\textsc{Else}}}

\newcommand{\DSLWhile}{\textcode{\textsc{While}}}
\newcommand{\DSLRun}{\textcode{\textsc{Run}}}
\newcommand{\DSLBoolGoal}{\textcode{goal}}
\newcommand{\DSLBoolPathAhead}{\textcode{pathAhead}}

\newcommand{\DSLBoolPathLeft}{\textcode{pathLeft}}

\newcommand{\DSLBoolPathRight}{\textcode{pathRight}}

\newcommand{\numEval}{\text{numEval}}
\newcommand{\GMove}{\textcode{gM}}
\newcommand{\GLeft}{\textcode{gL}}
\newcommand{\GRight}{\textcode{gR}}
\newcommand{\GStart}{\textcode{gStart}}
\newcommand{\GRepLeft}{\textcode{gRepL}}
\newcommand{\GRepRight}{\textcode{gRepR}}
\newcommand{\GRepMove}{\textcode{gRepM}}

\definecolor{TutorColour}{RGB}{105, 105, 105}
\definecolor{PCFGColour}{RGB}{204, 153, 24}

\newcommand{\student}{\textcode{stu}}

\newcommand{\qmark}{\textcode{?}}

\title{From \{Solution\} Synthesis to \{Student Attempt\} Synthesis for Block-Based Visual Programming Tasks\thanks{This article is a longer version of the paper from the EDM 2022 conference. Authors are listed alphabetically.}}

\numberofauthors{2}
\author{
\alignauthor
		Adish Singla\\
      \affaddr{MPI-SWS}\\
      \email{adishs@mpi-sws.org}
\alignauthor
	   Nikitas Theodoropoulos\\
      \affaddr{MPI-SWS}\\
      \email{ntheodor@mpi-sws.org}
}

\begin{document}
\maketitle

\newtoggle{longversion}
\settoggle{longversion}{false}

\begin{abstract}
Block-based visual programming environments are increasingly used to introduce computing concepts to beginners. Given that programming tasks are open-ended and conceptual, novice students often struggle when learning in these environments. AI-driven programming tutors hold great promise in automatically assisting struggling students, and need several components to realize this potential. We investigate the crucial component of student modeling, in particular, the ability to automatically infer students' misconceptions for predicting (synthesizing) their behavior. We introduce a novel benchmark, \benchmark, centered around the following challenge: For a given student, synthesize the student's attempt on a new target task after observing the student's attempt on a fixed reference task. This challenge is akin to that of program synthesis; however, instead of synthesizing a \emph{\{solution\}} (i.e., program an expert would write), the goal here is to synthesize a \emph{\{student attempt\}} (i.e., program that a given student would write). We first show that human experts (\algoTutor) can achieve high performance on the benchmark, whereas simple baselines perform poorly. Then, we develop two neuro/symbolic techniques (\algoContinual{} and \algoPCFG) in a quest to close this gap with \algoTutor.
\end{abstract}



\begin{figure*}[t!]
\centering
	\begin{subfigure}[b]{.40\textwidth}
	\centering
	{
		\begin{minipage}{0.29\textwidth}
			\includegraphics[height=2.3cm]{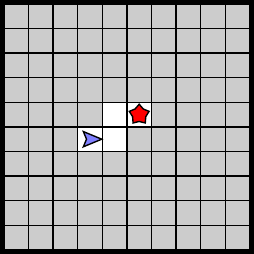}
		\end{minipage}
		\ 
		\begin{minipage}{0.44\textwidth}	
			\centering
			{
				\begin{boxcode}{3.4cm}{0.80}{0.8}
	                \textcode{def }\DSLRun\textcode{()}\textcode{\{}\\
    	            \quad\DSLMove\\
        	        \quad\DSLTurnLeft\\
            	    \quad\DSLMove\\
                	\quad\DSLTurnRight\\
                	\quad\DSLMove\\
                	\textcode{\}}\\
					\vspace{2.7mm}
				\end{boxcode}
    		}
		\end{minipage}
		\ 
		\begin{minipage}{0.20\textwidth}
			\includegraphics[height=2.1cm]{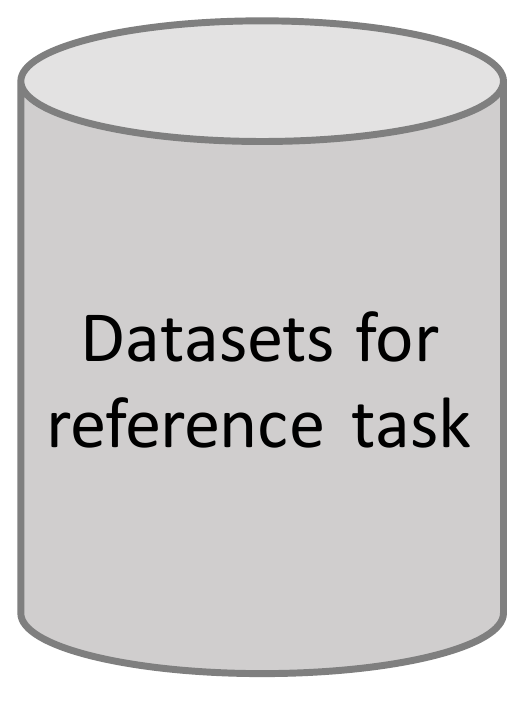}
		\end{minipage}
		\caption{Reference task $\task^{\textnormal{4}}$ with solution code and datasets}		
		 \label{fig:overview.hoc4.reftask}
	}
	\end{subfigure}
	\ \ 
	\begin{subfigure}[b]{.205\textwidth}
	\centering
	{
		\begin{boxcode}{3.4cm}{0.80}{0.8}
        	\textcode{def }\DSLRun\textcode{()}\textcode{\{}\\
        	\quad\DSLMove\\
        	\quad\DSLTurnLeft\\
        	\quad\DSLMove\\
        	\quad\DSLMove\\
        	\textcode{\}}\\
			\vspace{4.9mm}
		\end{boxcode}
		\vspace{-3mm}
		\caption{\student{}'s attempt for $\task^{\textnormal{4}}$}
		\label{fig:overview.hoc4.refcodestudent}
    }
    \end{subfigure}
	\begin{subfigure}[b]{.15\textwidth}
	\centering
	{
		\includegraphics[height=2.3cm]{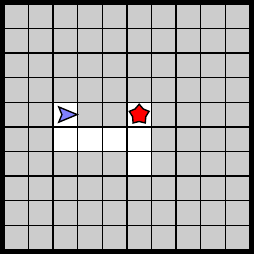}
		\caption{Target task $\task^{\textnormal{4x}}$}		
		 \label{fig:overview.hoc4.targettask}
	}
	\end{subfigure}
	\begin{subfigure}[b]{.21\textwidth}
	\centering
	{
		\begin{boxcode}{3.4cm}{0.80}{0.58}
			\\
			\\
			\\
			\vspace{2mm}			
			\hspace{1.2cm} {\fontsize{40}{1}\selectfont ?}
			\\
			\\
			\vspace{4mm}
		\end{boxcode}
		\vspace{-3mm}
		\caption{\student{}'s attempt for $\task^{\textnormal{4x}}$}		
		\label{fig:overview.hoc4.targetcodestudent}
    }
    \end{subfigure}
    \caption{\looseness-1Illustration of our problem setup and objective for the task Maze\#$4$ in the \emph{Hour of Code: Maze}~\cite{hourofcode_maze} by Code.org~\cite{codeorg}. As explained in Section~\ref{sec:problem.objectives}, we consider three distinct phases in our problem setup to provide a conceptual separation in terms of information and computation available to a system. \textbf{(a)} In the first phase, we are given a reference task $\task^{\textnormal{4}}$ along with its solution code $\code^{\star}_{\task^{\text{4}}}$ and data resources (e.g., a real-world dataset of different students' attempts); reference tasks are fixed and the system can use any computation a priori. \textbf{(b)} In the second phase, the system interacts with a student, namely \student{}, who attempts the reference task $\task^{\textnormal{4}}$ and submits a code, denoted as $\code^{\text{\student}}_{\task^{\textnormal{4}}}$. \textbf{(c, d)} In the third phase, the system seeks to synthesize the student \student{}'s behavior on a target task $\task^{\textnormal{4x}}$, i.e., a program that \student{} would write if the system would assign $\task^{\textnormal{4x}}$ to the student. Importantly, the target task $\task^{\textnormal{4x}}$ is not available a priori and this synthesis process would be done in real-time. Furthermore, the system may have to synthesize \student{}'s behavior on a large number of different target tasks (e.g., to personalize the next task in a curriculum).  Section~\ref{sec:problem} provides further details about the problem setup and objective; Section~\ref{sec:benchmark} introduces the \benchmark{} benchmark comprising of different types of students and target tasks for the reference task.}
	\label{fig:intro.hoc4}
\end{figure*}
%
\begin{figure*}[t!]
\centering
	\begin{subfigure}[b]{.40\textwidth}
	\centering
	{
		\begin{minipage}{0.29\textwidth}
			\includegraphics[height=2.3cm]{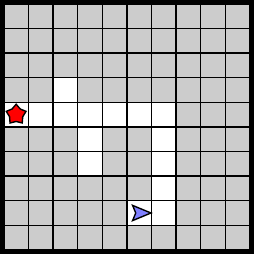}
		\end{minipage}
		\ 
		\begin{minipage}{0.44\textwidth}	
			\centering
			{
				\begin{boxcode}{3.4cm}{0.80}{0.54}
	                \textcode{def }\DSLRun\textcode{()}\textcode{\{}\\
    	            \quad\DSLRepeatUntil\textcode{(\DSLBoolGoal)}\textcode{\{}\\
        	        \quad\quad\DSLIf\textcode{(\DSLBoolPathAhead)}\textcode{\{}\\
            	    \quad\quad\quad\DSLMove\\\quad\quad\textcode{\}}\\
            	    \quad\quad\DSLElse\textcode{\{}\\
                	\quad\quad\quad\DSLTurnLeft\\
                	\quad\quad\textcode{\}}\\
                	\quad\textcode{\}}\\
                	\textcode{\}}
				\end{boxcode}
    		}
		\end{minipage}	
		\ 
		\begin{minipage}{0.20\textwidth}
			\includegraphics[height=2.1cm]{fig/_intro/data.pdf}
		\end{minipage}
		\
		\caption{Reference task $\task^{\textnormal{18}}$ with solution code and datasets}				
		 \label{fig:overview.hoc18.reftask}
	}
	\end{subfigure}
	\ \ 	
	\begin{subfigure}[b]{.205\textwidth}
	\centering
	{
		\begin{boxcode}{3.4cm}{0.80}{0.58}
				\textcode{def }\DSLRun\textcode{()}\textcode{\{}\\
                \quad\DSLRepeatUntil\textcode{(\DSLBoolGoal)}\textcode{\{}\\
               	    \quad\quad\DSLMove\\
                    \quad\quad\DSLTurnLeft\\
               	    \quad\quad\DSLMove\\
                    \quad\quad\DSLTurnLeft\\
               	    \quad\quad\DSLMove\\
                \quad\textcode{\}}\\
                \textcode{\}}
			     \vspace{3.3mm}
		\end{boxcode}
		\vspace{-3mm}
		\caption{\student{}'s attempt for $\task^{\textnormal{18}}$}
		\label{fig:overview.hoc18.refcodestudent}
    }
    \end{subfigure}
	\begin{subfigure}[b]{.15\textwidth}
	\centering
	{
		\includegraphics[height=2.3cm]{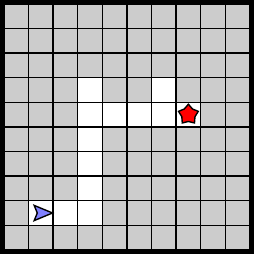}
		\caption{Target task $\task^{\textnormal{18x}}$}		
		 \label{fig:overview.hoc18.targettask}
	}
	\end{subfigure}
	\begin{subfigure}[b]{.21\textwidth}
	\centering
	{
		\begin{boxcode}{3.4cm}{0.80}{0.58}
			\\
			\\
			\\
			\vspace{2mm}			
			\hspace{1.2cm} {\fontsize{40}{1}\selectfont ?}
			\\
			\\
			\vspace{4mm}
		\end{boxcode}
		\vspace{-3mm}
		\caption{\student{}'s attempt for $\task^{\textnormal{18x}}$}		
		\label{fig:overview.hoc18.targetcodestudent}
    }
    \end{subfigure}
    \caption{Analogous to Figure~\ref{fig:intro.hoc4}, here we illustrate the setup for the task Maze\#$18$ in the \emph{Hour of Code: Maze Challenge}~\cite{hourofcode_maze}.
    }
	\vspace{-1mm}	
	\label{fig:intro.hoc18}
\end{figure*}

\section{Introduction}\label{sec:intro}
The emergence of block-based visual programming platforms has made coding more accessible and appealing to beginners. Block-based programming uses ``code blocks'' that reduce the burden of syntax and introduces concepts in an  interactive way. Led by initiatives like \emph{Hour of Code} by Code.org~\cite{hourofcode,codeorg} and the popularity of languages like \emph{Scratch}~\cite{DBLP:journals/cacm/ResnickMMREBMRSSK09}, block-based programming has become integral to introductory CS education. Considering the \emph{Hour of Code} initiative alone, over one billion hours of programming activity has been spent in learning to solve tasks in such environments~\cite{codeorg}.

Programming tasks on these platforms are conceptual and open-ended, and require multi-step deductive reasoning to solve. Given these aspects, novices often struggle when learning to solve these tasks. The difficulties faced by novice students become evident by looking at the trajectory of students' attempts who are struggling to solve a given task. For instance, in a dataset released by Code.org~\cite{hourofcode,codeorg,DBLP:conf/lats/PiechSHG15}, even for simple tasks where solutions require only $5$ code blocks (see Figure~\ref{fig:overview.hoc18.reftask}), students submitted over $50,000$ unique attempts with some exceeding a size of $50$ code blocks. 

AI-driven programming tutors have the potential to support these struggling students by providing personalized assistance, e.g., feedback as hints or curriculum design~\cite{price2017position}. To effectively assist struggling students, AI-driven systems need several components, a crucial one being \emph{student modeling}. In particular, we need models that can automatically infer a student's knowledge from limited interactions and then predict the student's behavior on new tasks. However, student modeling in block-based visual programming environments can be quite challenging because of the following: (i) programming tasks are conceptual, and there is no well-defined skill-set or problem-solving strategy for mastery~\cite{kaser2020modeling}; (ii) there could be a huge variability in behaviors and a long-tail distribution of students' attempts for a task~\cite{DBLP:conf/aaai/WuMGP19}; (iii) the objective of predicting a student's behavior on new tasks is not limited to coarse-grained success/failure indicators (e.g., \cite{Wang2017LearningTR})---ideally, we should be able to do fine-grained synthesis of attempts for a given student.

Beyond the above-mentioned challenges, there are two critical issues arising from limited resources and data scarcity for a given domain. First, while the space of tasks that could be designed for personalized curriculum is intractably large~\cite{DBLP:conf/nips/AhmedCEFGRS20}, the publicly available datasets of real-world students' attempts are limited; e.g., for the \emph{Hour of Code: Maze Challenge} domain, we have datasets for only two tasks~\cite{DBLP:conf/lats/PiechSHG15}. Second, when a deployed system is interacting with a new student, there is limited prior information~\cite{DBLP:conf/edm/EfremovGS20}, and the system would have to infer the student's knowledge by observing behavior on a few reference tasks, e.g., through a quiz~\cite{DBLP:conf/edm/He-YueyaS21}. These two issues, in turn, limit the applicability of state-of-the-art techniques that rely on large-scale datasets across tasks or personalized data per student (e.g., \cite{Wang2017LearningTR,DBLP:conf/kdd/McIlroy-Young0K20,McIlroyYoung2021DetectingID,PortnoffMethodsFL})---we need next-generation student modeling techniques for block-based visual programming that can operate under data scarcity and limited observability. To this end, this paper focuses on the following question:
\vspace{-4mm}
\begin{quote}
\emph{For a given student, can we synthesize the student's attempt on a new target task after observing the student's attempt on a fixed reference task?}
\end{quote}
\vspace{-2mm}

\subsection{Our Approach and Contributions}\label{sec:intro.contributions}
Figures~\ref{fig:intro.hoc4}~and~\ref{fig:intro.hoc18} illustrate this synthesis question for two scenarios in the context of the \emph{Hour of Code: Maze Challenge}~\cite{hourofcode_maze} by Code.org~\cite{codeorg}. This question is akin to that of program synthesis~\cite{gulwani2017program}; however, instead of synthesizing a \emph{\{solution\}} (i.e., program an expert would write), the goal here is to synthesize a \emph{\{student attempt\}} (i.e., program that a given student would write). This goal of synthesizing student attempts, and not just solutions, requires going beyond state-of-the-art program synthesis techniques~\cite{DBLP:conf/iclr/BunelHDSK18, DBLP:conf/iclr/ChenLS19,li_competition-level}; crucially, we also need to define appropriate metrics to quantitatively measure the performance of different techniques. Our approach and contributions are summarized below:
\vspace{-3mm}
\setlength{\leftmargini}{1.5em}
\begin{enumerate}[label=(\arabic*),parsep=1pt]
\item We formalize the problem of synthesizing a student's attempt on target tasks after observing the student's behavior on a fixed reference task. We introduce a novel benchmark, \benchmark, centered around the above synthesis question, along with generative/discriminative performance measures for evaluation.  (Sections~\ref{sec:problem},~\ref{sec:benchmark.data},~\ref{sec:benchmark.measures})
\vspace{-3mm}
\item We showcase that human experts (\algoTutor{}) can achieve high performance on \benchmark, whereas  simple baselines perform poorly.  (Section~\ref{sec:benchmark.initialresults})
\item We develop two techniques inspired by neural (\algoContinual{}) and symbolic (\algoPCFG{}) methods, in a quest to close the gap with human experts (\algoTutor{}).  (Sections~\ref{sec:continual},~\ref{sec:pcfg},~\ref{sec:experiments})
\item We publicly release the benchmark and implementations to facilitate future research.\footnote{The \benchmark{} benchmark and implementation of the techniques are available at \url{https://github.com/machine-teaching-group/edm2022\_studentsyn}.}
%
\end{enumerate}
\setlength{\leftmargini}{2.5em}


\subsection{Related Work}\label{sec:intro.related}
\textbf{Student modeling.}  Inferring the knowledge state of a student is an integral part of AI tutoring systems and relevant to our goal of predicting a student's behavior. For close-ended domains like vocabulary learning (\cite{DBLP:conf/acl/SettlesM16,PortnoffMethodsFL,DBLP:conf/nips/Hunziker0AR0PYS19}) and Algebra problems (\cite{DBLP:journals/umuai/CorbettMS00, DBLP:conf/edm/RaffertyJG16, shakya_student}), the  skills or knowledge components for mastery are typically well-defined and we can use \emph{Knowledge Tracing} techniques to model a student's knowledge state over time \cite{corbett1994knowledge,DBLP:conf/nips/PiechBHGSGS15}. These modeling techniques, in turn, allow us to provide feedback, predict solution strategies, or infer/quiz a student's knowledge state~\cite{DBLP:conf/edm/RaffertyJG16,DBLP:conf/edm/He-YueyaS21,shakya_student}. Open-ended domains pose unique challenges to directly apply these techniques (see~\cite{kaser2020modeling}); however, there has been some progress in this direction. In recent works \cite{DBLP:conf/kdd/McIlroy-Young0K20,McIlroyYoung2021DetectingID}, models have been proposed to predict human behavior in chess for specific skill levels and to recognize the behavior of individual players. Along these lines, \cite{DBLP:conf/edm/CockMGK21} introduced methods to perform early prediction of struggling students in open-ended interactive simulations. There has also been work on student modeling for block-based programming, e.g., clustering-based methods for misconception discovery~\cite{DBLP:conf/sigcse/EmersonSRWMBL20,DBLP:conf/lak/0004SWMPP21}, and deep learning methods to represent knowledge and predict future performance~\cite{Wang2017LearningTR}.

\textbf{AI-driven systems for programming education.} There has been a surge of interest in developing AI-driven systems for programming education, and in particular, for block-based programming domains~\cite{price2017position,price2017isnap,weintrop2017comparing}. Existing works have studied various aspects of intelligent feedback, for instance, providing next-step hints when a student is stuck~\cite{DBLP:conf/lats/PiechSHG15,yi2017feasibility,DBLP:journals/corr/abs-1708-06564,DBLP:conf/edm/EfremovGS20}, giving data-driven feedback about a student's misconceptions~\cite{singh2013automated,DBLP:conf/icml/PiechHNPSG15,price2017evaluation,DBLP:conf/aaai/WuMGP19}, or generating/recommending new tasks~\cite{DBLP:conf/edm/AiCGZWFW19,DBLP:conf/nips/AhmedCEFGRS20,aied21_popquiz}. Depending on the availability of datasets and resources, different techniques are employed: using historical datasets to learn code embeddings \cite{DBLP:conf/icml/PiechHNPSG15,DBLP:journals/corr/abs-1708-06564}, using reinforcement learning in zero-shot setting~\cite{DBLP:conf/edm/EfremovGS20,DBLP:journals/corr/abs-2107-08828}, bootstrapping from a small set of expert annotations~\cite{DBLP:conf/icml/PiechHNPSG15}, or using expert grammars to generate synthetic training data~\cite{DBLP:conf/aaai/WuMGP19}.
%

\looseness-1\textbf{Neuro-symbolic program synthesis.} 
Our approach is related to program synthesis, i.e., automatically constructing programs that satisfy a given specification~\cite{gulwani2017program}. In recent years, the usage of deep learning models for program synthesis has resulted in significant progress in a variety of domains including string transformations \cite{DBLP:conf/nips/EllisNPSTS19,DBLP:conf/icml/DevlinUBSMK17,DBLP:conf/iclr/ParisottoMS0ZK17}, block-based visual programming~\cite{DBLP:conf/iclr/BunelHDSK18, DBLP:conf/iclr/ChenLS19, DBLP:conf/nips/DevlinBSHK17,DBLP:journals/corr/abs-2108-13643}, and competitive programming~\cite{li_competition-level}. Program synthesis has also been used to learn compositional symbolic rules and mimic abstract human learning \cite{DBLP:conf/nips/NyeS0L20,DBLP:journals/corr/abs-2006-08381}.  Our goal is akin to program synthesis and we leverage the work of \cite{DBLP:conf/iclr/BunelHDSK18} in our technique \algoContinual, however, with a crucial difference: instead of synthesizing a solution program, we seek to synthesize a student's attempt.
%

\section{Problem Setup}\label{sec:problem}
Next, we introduce definitions and formalize our objective.
%

\subsection{Preliminaries}\label{sec:problem.preliminaries}
\textbf{The space of tasks.} We define the space of tasks as~\taskspace; in this paper, \taskspace{} is inspired by the popular \emph{Hour of Code: Maze Challenge}~\cite{hourofcode_maze} from Code.org~\cite{codeorg}; see Figures~\ref{fig:overview.hoc4.reftask}~and~\ref{fig:overview.hoc18.reftask}. We define a task $\task \in \taskspace$ as a tuple $(\task_\textnormal{vis}, \task_\textnormal{store}, \task_\textnormal{size})$, where $\task_\textnormal{vis}$ denotes a visual puzzle,  $\task_\textnormal{store}$ the available block types, and $\task_\textnormal{size}$ the maximum number of blocks allowed in the solution code.  For instance, considering the task $\task$ in Figure~\ref{fig:overview.hoc18.reftask}, we have the following specification: the visual puzzle $\task_\textnormal{vis}$ comprises of a maze where the objective is to navigate the ``avatar'' (blue-colored triangle) to the ``goal'' (red-colored star) by executing a code; the set of available types of blocks $\task_\textnormal{store}$ is \{\DSLMove, \DSLTurnLeft, \DSLTurnRight, \DSLRepeatUntil{(\DSLBoolGoal}), \DSLIfElse{(\DSLBoolPathAhead}), \DSLIfElse{(\DSLBoolPathLeft}), \DSLIfElse{(\DSLBoolPathRight})\}, and the size threshold $\task_\textnormal{size}$ is $5$ blocks; this particular task in Figure~\ref{fig:overview.hoc18.reftask} corresponds to Maze\#$18$ in the \emph{Hour of Code: Maze Challenge}~\cite{hourofcode_maze}, and has been studied in a number of prior works~\cite{DBLP:conf/lats/PiechSHG15,DBLP:conf/edm/EfremovGS20,DBLP:conf/nips/AhmedCEFGRS20}.
%

%
\textbf{The space of codes.\footnote{Codes are also interchangeably referred to as programs.}} We define the space of all possible codes as \codespace{} and represent them using a \emph{Domain Specific Language }(DSL)~\cite{gulwani2017program}. In particular, for codes relevant to  tasks considered in this paper, we use a DSL from~\cite{DBLP:conf/nips/AhmedCEFGRS20}. A code $\code \in \codespace$ has the following attributes: $\code_\text{blocks}$ is the set of types of code blocks used in $\code$, $\code_\text{size}$ is the number of code blocks used, and $\code_\text{depth}$ is the depth of the \emph{Abstract Syntax Tree} of $\code$. Details of this DSL and code attributes are not crucial for the readability of subsequent sections; however, they provide useful formalism when implementing different techniques introduced in this paper.

%
\textbf{Solution code and student attempt.} For a given task $\task$, a \emph{solution code} $\code^{\star}_{\task} \in \codespace$ should solve the visual puzzle; additionally, it can only use the allowed types of code blocks (i.e., $\code_\text{blocks} \subseteq \task_\textnormal{store}$) and should be within the specified size threshold (i.e., $\code_\text{size} \leq \task_\textnormal{size}$). We note that a task $\task \in \taskspace$ in general may have multiple solution codes; in this paper, we typically refer to a single solution code that is provided as input. A \emph{student attempt} for a task $\task$ refers to a code that is written by a student (including incorrect or partial codes). A student attempt could be any code $\code \in \codespace$ as long as it uses the set of available types of code blocks (i.e., $\code_\text{blocks} \subseteq \task_\textnormal{store}$); importantly, it is not restricted by the size threshold $\task_\textnormal{size}$---same setting as in the programming environment of \emph{Hour of Code: Maze Challenge}~\cite{hourofcode_maze}.

\begin{figure*}[t!]
\centering
		\begin{subfigure}[b]{0.40\textwidth}
		\centering
		{
			\begin{minipage}{0.29\textwidth}
				\includegraphics[height=2.3cm]{fig/_tasks/hoc4_task.pdf}
			\end{minipage}
			\ 
			\begin{minipage}{0.44\textwidth}	
			\centering
			{
				\begin{boxcode}{3.4cm}{0.80}{0.58}
	               	\textcode{def }\DSLRun\textcode{()}\textcode{\{}\\
    	           	\quad\DSLMove\\
        	       	\quad\DSLTurnLeft\\
            	   	\quad\DSLMove\\
                	\quad\DSLTurnRight\\
                	\quad\DSLMove\\
                	\textcode{\}}\\
					\vspace{7.3mm}
				\end{boxcode}
    		}
			\end{minipage}
			\ 
			\begin{minipage}{0.20\textwidth}
				\includegraphics[height=2.1cm]{fig/_intro/data.pdf}
			\end{minipage}
			\caption{Reference task $\task^{\textnormal{4}}$ with solution code and datasets}		
		 	\label{fig:benchmark.hoc4.reftask}
		}
		\end{subfigure}	
		\quad \quad
		\begin{subfigure}[b]{.55\textwidth}
		\centering
		{
		    \ \ \ \ 
			\begin{minipage}{0.30\textwidth}
				\includegraphics[height=2.3cm]{fig/_tasks/hoc4a_task.pdf}
			\end{minipage}
			\begin{minipage}{0.30\textwidth}
				\includegraphics[height=2.3cm]{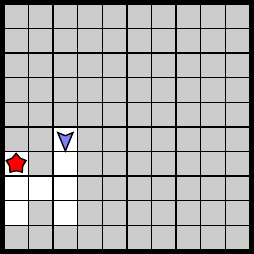}
			\end{minipage}
			\begin{minipage}{0.30\textwidth}
				\includegraphics[height=2.3cm]{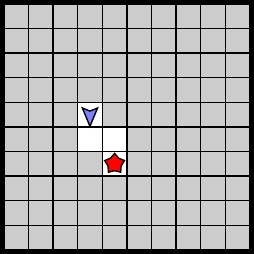}
			\end{minipage}
			\caption{Three target tasks for $\task^{\textnormal{4}}$: $\task^{\textnormal{4x}}$, $\task^{\textnormal{4y}}$, and $\task^{\textnormal{4z}}$}
		 	\label{fig:benchmark.hoc4.targettasks}			
		}
		\end{subfigure}
		\begin{subfigure}[b]{1\textwidth}
		\vspace{2mm}
		\centering
		{
			\begin{minipage}{0.15\textwidth}	
			\centering
			{
				\begin{boxcode}{3cm}{0.80}{0.80}
	 				\textcode{def }\DSLRun\textcode{()}\textcode{\{}\\
                	\quad\DSLMove\\
                	\quad\DSLTurnRight\\
               	 	\quad\DSLMove\\
                	\quad\DSLTurnLeft\\
                	\quad\DSLMove\\
                	\textcode{\}}\\
				 	\vspace{8.4mm}
				\end{boxcode}
    		}
    		\end{minipage}
    		\
			\begin{minipage}{0.15\textwidth}	
			\centering
			{
				\begin{boxcode}{3cm}{0.80}{0.80}
        			\textcode{def }\DSLRun\textcode{()}\textcode{\{}\\
                	\quad\DSLMove\\
               	 	\quad\DSLTurnLeft\\
                	\quad\DSLMove\\
                	\textcode{\}}\\
					\vspace{14.2mm}
				\end{boxcode}
    		}
    		\end{minipage}
    		\
			\begin{minipage}{0.15\textwidth}	
			\centering
			{
				\begin{boxcode}{3cm}{0.80}{0.79}
        			\textcode{def }\DSLRun\textcode{()}\textcode{\{}\\
               	 	\quad\DSLMove\\
               	 	\quad\DSLTurnRight\\
                	\quad\DSLTurnLeft\\
                	\quad\DSLTurnRight\\
                	\quad\DSLMove\\
                	\quad\DSLTurnLeft\\
                	\quad\DSLTurnLeft\\
                	\quad\DSLTurnRight\\
                	\quad\DSLMove\\
                	\textcode{\}}
				\end{boxcode}
    		}
    		\end{minipage}
    		\    		
			\begin{minipage}{0.15\textwidth}	
			\centering
			{
				\begin{boxcode}{3cm}{0.80}{0.71}
        			\textcode{def }\DSLRun\textcode{()}\textcode{\{}\\
              	  	\quad\DSLMove\\
                	\quad\DSLMove\\
                	\quad\DSLMove\\
                	\quad\DSLTurnLeft\\
                	\quad\DSLMove\\
                	\quad\DSLMove\\
                	\quad\DSLMove\\
                	\quad\DSLMove\\
                	\quad\DSLTurnRight\\
                	\quad\DSLMove\\
                	\textcode{\}}
					\vspace{0.3mm}
				\end{boxcode}
    		}
    		\end{minipage}
    		\    		
			\begin{minipage}{0.15\textwidth}	
			\centering
			{
				\begin{boxcode}{3cm}{0.80}{0.80}
        			\textcode{def }\DSLRun\textcode{()}\textcode{\{}\\
        			\quad\DSLMove\\
        			\quad\DSLTurnLeft\\
        			\quad\DSLMove\\
        			\quad\DSLMove\\
             	   \textcode{\}}\\
					\vspace{11.6mm}
				\end{boxcode}
    		}
    		\end{minipage}
    		\    		
			\begin{minipage}{.19\textwidth}	
			\centering
			{
				\begin{boxcode}{3.35cm}{0.80}{0.54}
        			\textcode{def }\DSLRun\textcode{()}\textcode{\{}\\
                	\quad\DSLMove\\
               	 	\quad\DSLMove\\
                	\quad\DSLTurnLeft\\
                	\quad\DSLTurnRight\\
                	\quad\DSLTurnRight\\
                	\quad\DSLTurnLeft\\
               	 	\quad\DSLMove\\
                	\quad\DSLTurnLeft\\
                	\quad\DSLTurnRight\\
                	\quad\textcode{...}\\
                	\vspace{0.1mm}
                	\quad\textcode{(many more blocks) }\\
                	\textcode{\}}
				\end{boxcode}
    		}
    		\end{minipage}								
			\caption{Example codes (i)--(vi) corresponding to six types of students' behaviors when attempting $\task^{\textnormal{4}}$, each capturing different misconceptions}
			\label{fig:benchmark.hoc4.students}
		}
		\end{subfigure}
        \caption{Illustration of the key elements of the \benchmark{} benchmark for the reference task $\task^{\textnormal{4}}$ shown in \textbf{(a)}---same as in Figure~\ref{fig:overview.hoc4.reftask}. \textbf{(b)} Shows three target tasks associated with $\task^{\textnormal{4}}$; these target tasks are similar to $\task^{\textnormal{4}}$ in a sense that the set of available block types is same as $\task^{4}_\textnormal{store}$ and the nesting structure of programming constructs in solution codes is same as in $\code^{\star}_{\task^{\textnormal{4}}}$. \textbf{(c)} Shows example codes corresponding to six types of students' behaviors when attempting $\task^{\textnormal{4}}$, each capturing a different misconception as follows: (i) confusing left/right directions when turning, (ii) partially solving the task in terms of getting closer to the ``goal'', (iii) misunderstanding of turning functionality and writing repetitive turn commands, (iv) adding more than the correct number of required move commands, (v) forgetting to include some turns needed in the solution,  (vi) attempting to randomly solve the task by adding lots of blocks. See details in Section~\ref{sec:benchmark.data}.
        }
		\label{fig:benchmark.hoc4}		
\end{figure*}

%
\begin{figure*}[t!]
\centering
		\begin{subfigure}[b]{0.40\textwidth}
		\centering
		{
			\begin{minipage}{0.29\textwidth}
				\includegraphics[height=2.3cm]{fig/_tasks/hoc18_task.pdf}
			\end{minipage}
			\ 
			\begin{minipage}{0.44\textwidth}	
			\centering
			{
				\begin{boxcode}{3.4cm}{0.80}{0.54}
	                \textcode{def }\DSLRun\textcode{()}\textcode{\{}\\
    	            \quad\DSLRepeatUntil\textcode{(\DSLBoolGoal)}\textcode{\{}\\
        	        \quad\quad\DSLIf\textcode{(\DSLBoolPathAhead)}\textcode{\{}\\
            	    \quad\quad\quad\DSLMove\\\quad\quad\textcode{\}}\\
            	    \quad\quad\DSLElse\textcode{\{}\\
                	\quad\quad\quad\DSLTurnLeft\\
                	\quad\quad\textcode{\}}\\
                	\quad\textcode{\}}\\
                	\textcode{\}}
				\end{boxcode}
    		}
			\end{minipage}
			\ 
			\begin{minipage}{0.20\textwidth}
				\includegraphics[height=2.1cm]{fig/_intro/data.pdf}
			\end{minipage}
			\caption{Reference task $\task^{\textnormal{18}}$ with solution code and datasets}		
		 	\label{fig:benchmark.hoc18.reftask}
		}
		\end{subfigure}	
		\quad \quad
		\begin{subfigure}[b]{.55\textwidth}
		\centering
		{
		    \ \ \ \ 
			\begin{minipage}{0.30\textwidth}
				\includegraphics[height=2.3cm]{fig/_tasks/hoc18b_task.pdf}
			\end{minipage}
			\begin{minipage}{0.30\textwidth}
				\includegraphics[height=2.3cm]{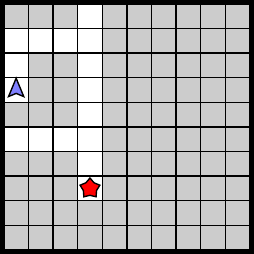}
			\end{minipage}
			\begin{minipage}{0.30\textwidth}
				\includegraphics[height=2.3cm]{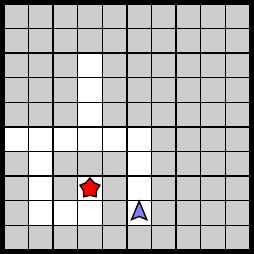}
			\end{minipage}
			\caption{Three target tasks for $\task^{\textnormal{18}}$: $\task^{\textnormal{18x}}$, $\task^{\textnormal{18y}}$, and $\task^{\textnormal{18z}}$}
		 	\label{fig:benchmark.hoc18.targettasks}			
		}
		\end{subfigure}
		\begin{subfigure}[b]{1\textwidth}
		\vspace{2mm}
		{
			\begin{minipage}{0.155\textwidth}	
			\centering
			{
				\begin{boxcode}{3.4cm}{0.75}{0.80}
	 				\textcode{def }\DSLRun\textcode{()}\textcode{\{}\\
                	\quad\DSLRepeatUntil\textcode{(\DSLBoolGoal)}\textcode{\{}\\
               	 	\quad\quad\DSLIf\textcode{(\DSLBoolPathAhead)}\textcode{\{}\\
                	\quad\quad\quad\DSLMove\\
                	\quad\quad\textcode{\}}\\
                	\quad\quad\DSLElse\textcode{\{}\\
                	\quad\quad\quad\DSLTurnRight\\
                	\quad\quad\textcode{\}}\\
                	\quad\textcode{\}}\\
                	\textcode{\}}
				 	\vspace{4.0mm}
				\end{boxcode}
    		}
    		\end{minipage}
    		\
			\begin{minipage}{0.155\textwidth}	
			\centering
			{
				\begin{boxcode}{3.4cm}{0.75}{0.80}
        			\textcode{def }\DSLRun\textcode{()}\textcode{\{}\\
                	\quad\DSLRepeatUntil\textcode{(\DSLBoolGoal)}\textcode{\{}\\
               	 	\quad\quad\DSLIf\textcode{(\DSLBoolPathLeft)}\textcode{\{}\\
                	\quad\quad\quad\DSLTurnLeft\\
                	\quad\quad\quad\DSLMove\\
                	\quad\quad\textcode{\}}\\
                	\quad\quad\DSLElse\textcode{\{}\\
                	\quad\quad\quad\DSLMove\\
                	\quad\quad\textcode{\}}\\
                	\quad\textcode{\}}\\
                	\textcode{\}}					\vspace{1.0mm}
				\end{boxcode}				
    		}
    		\end{minipage}
    		\
			\begin{minipage}{0.155\textwidth}	
			\centering
			{
				\begin{boxcode}{3.4cm}{0.75}{0.80}
        			\textcode{def }\DSLRun\textcode{()}\textcode{\{}\\
                	\quad\DSLRepeatUntil\textcode{(\DSLBoolGoal)}\textcode{\{}\\
                	\quad\quad\DSLIf\textcode{(\DSLBoolPathAhead)}\textcode{\{}\\
                	\quad\quad\quad\DSLTurnLeft\\
                	\quad\quad\textcode{\}}\\
                	\quad\quad\DSLElse\textcode{\{}\\
               	 	\quad\quad\quad\DSLTurnLeft\\
                	\quad\quad\textcode{\}}\\
                	\quad\quad\DSLMove\\
                	\quad\textcode{\}}\\
                	\textcode{\}}
                	\vspace{1.1mm}
				\end{boxcode}
    		}
    		\end{minipage}
    		\
			\begin{minipage}{0.155\textwidth}	
			\centering
			{
				\begin{boxcode}{3.4cm}{0.75}{0.80}
        			\textcode{def }\DSLRun\textcode{()}\textcode{\{}\\
                	\quad\DSLRepeatUntil\textcode{(\DSLBoolGoal)}\textcode{\{}\\
                	\quad\quad\DSLMove\\
                    \quad\quad\DSLTurnLeft\\
               	    \quad\quad\DSLMove\\
                    \quad\quad\DSLTurnLeft\\
               	    \quad\quad\DSLMove\\
                \quad\textcode{\}}\\
                \textcode{\}}
			     \vspace{7.6mm}
				\end{boxcode}
    		}
    		\end{minipage}
    		\
			\begin{minipage}{0.155\textwidth}	
			\centering
			{
				\begin{boxcode}{3.4cm}{0.75}{0.80}
                    \textcode{def }\DSLRun\textcode{()}\textcode{\{}\\
                	\quad\DSLMove\\
                	\quad\DSLIf\textcode{(\DSLBoolPathAhead)}\textcode{\{}\\
                	\quad\quad\DSLMove\\
                	\quad\textcode{\}}\\
                	\quad\DSLElse\textcode{\{}\\
                	\quad\quad\DSLTurnLeft\\
                	\quad\textcode{\}}\\
                	\textcode{\}}\\
					\vspace{4.8mm}
				\end{boxcode}
    		}
    		\end{minipage}
    		\
			\begin{minipage}{0.155\textwidth}	
			\centering
			{
				\begin{boxcode}{3.2cm}{0.75}{0.66}
        			\textcode{def }\DSLRun\textcode{()}\textcode{\{}\\
                	\quad\DSLMove\\
                	\quad\DSLTurnLeft\\
                	\quad\DSLMove\\
                	\quad\DSLMove\\
                	\quad\DSLMove\\
                	\quad\DSLMove\\
                	\quad\DSLTurnRight\\
                	\quad\DSLMove\\
                	\quad\DSLMove\\
                	\quad\DSLMove\\
                	\quad\DSLMove\\
                	\quad\DSLMove\\
                	\textcode{\}}
				\end{boxcode}
    		}
    		\end{minipage}								
			\caption{Example codes (i)--(vi) corresponding to six types of students' behaviors when attempting $\task^{\textnormal{18}}$, each capturing different misconceptions}
			\label{fig:benchmark.hoc18.students}
		}
		\end{subfigure}
        \caption{Analogous to Figure~\ref{fig:benchmark.hoc4}, here we illustrate the key elements of the \benchmark{} benchmark for the reference task $\task^{\textnormal{18}}$ shown in \textbf{(a)}---same as in Figure~\ref{fig:overview.hoc18.reftask}. \textbf{(b)} Shows three target tasks associated with $\task^{\textnormal{18}}$. \textbf{(c)} Shows example codes corresponding to six types of students' behaviors when attempting $\task^{\textnormal{18}}$, each capturing a different misconception as follows: (i) confusing left/right directions when turning or checking conditionals, (ii) following one of the wrong path segments, 
        (iii) misunderstanding of \DSLIfElse{} structure functionality and writing the same blocks in both the execution branches, (iv) ignoring the \DSLIfElse{} structure when solving the task, (v) ignoring the \DSLWhile{} structure when solving the task,  (vi) attempting to solve the task by using only the 
        basic action blocks in \{\DSLTurnLeft, \DSLTurnRight, \DSLMove{}\}. See details in Section~\ref{sec:benchmark.data}.
        }
		\label{fig:benchmark.hoc18}	
	    \vspace{-2mm}		
\end{figure*}

%

\subsection{Objective}\label{sec:problem.objectives}

\textbf{Distinct phases.}  To formalize our objective, we introduce three distinct phases in our problem setup that provide a conceptual separation in terms of information and computation available to a system. More concretely, we have:

\vspace{-3mm}
\setlength{\leftmargini}{1.3em}
\begin{enumerate}[label=(\arabic*)] 
    \item Reference task $\task^{\text{ref}}$: We are given a reference task $\task^{\text{ref}}$ for which we have real-world datasets of different students' attempts as well as access to other data resources. Reference tasks are fixed and the system can use any computation a priori (e.g., compute code embeddings).
    \item Student \student{} attempts $\task^{\text{ref}}$: The system interacts with a student, namely \student{}, who attempts the reference task $\task^{\text{ref}}$ and submits a code, denoted as $\code^{\text{\student}}_{\task^{\text{ref}}}$. At the end of this phase, the system has observed \student{}'s behavior on $\task^{\text{ref}}$ and we denote this observation by the tuple $(\task^{\text{ref}}, \code^{\text{\student}}_{\task^{\text{ref}}})$.\footnote{In practice, the system might have more information, e.g., the whole trajectory of edits leading to $\code^{\text{\student}}_{\task^{\text{ref}}}$ or access to some prior information about the student \student{}.}
    \item Target task $\task^{\text{tar}}$: The system seeks to synthesize the student \student{}'s behavior on a target task $\task^{\text{tar}}$. Importantly, the target task $\task^{\text{tar}}$ is not available a priori and this synthesis process would be done in real-time, possibly with constrained computational resources. Furthermore, the system may have to synthesize \student{}'s behavior on a large number of different target tasks from the space \taskspace{} (e.g., to personalize the next task in a curriculum).\footnote{Even though the \emph{Hour of Code: Maze Challenge}~\cite{hourofcode_maze} has only $20$ tasks, the space \taskspace{} is intractably large and new tasks can be generated automatically, e.g., when providing feedback or for additional practice~\cite{DBLP:conf/nips/AhmedCEFGRS20}.} 
    \vspace{-3mm}
\end{enumerate}
\setlength{\leftmargini}{2.5em}

\looseness-1\textbf{Granularity level of our objective.}  
There are several different granularity levels at which we can predict the student \student{}'s behavior for $\task^{\text{tar}}$, including: (a) a coarse-level binary prediction of whether \student{} will successfully solve $\task^{\text{tar}}$, (b) a medium-level prediction about \student{}'s behavior w.r.t. a predefined feature set (e.g., labelled misconceptions); (c) a fine-level prediction in terms of synthesizing $\code^{\text{\student}}_{\task^{\text{tar}}}$, i.e., a program that \student{} would write if the system would assign $\task^{\text{tar}}$ to the student. In this work, we focus on this fine-level, arguably also the most challenging, synthesis objective.

\looseness-1\textbf{Performance evaluation.} So far, we have concretized the synthesis objective; however, there is still a question of how to quantitatively measure the performance of a technique set out to achieve this objective. The key challenge stems from the open-ended and conceptual nature of programming tasks. Even for seemingly simple tasks such as in Figures~\ref{fig:overview.hoc4.reftask}~and~\ref{fig:overview.hoc18.reftask}, the students' attempts can be highly diverse, thereby making it difficult to detect a student's misconceptions from observed behaviors; moreover, the space of misconceptions itself is not clearly understood. To this end, we begin by designing a benchmark to quantitatively measure  the performance of different techniques w.r.t. our objective.
%

\begin{figure*}[t!]
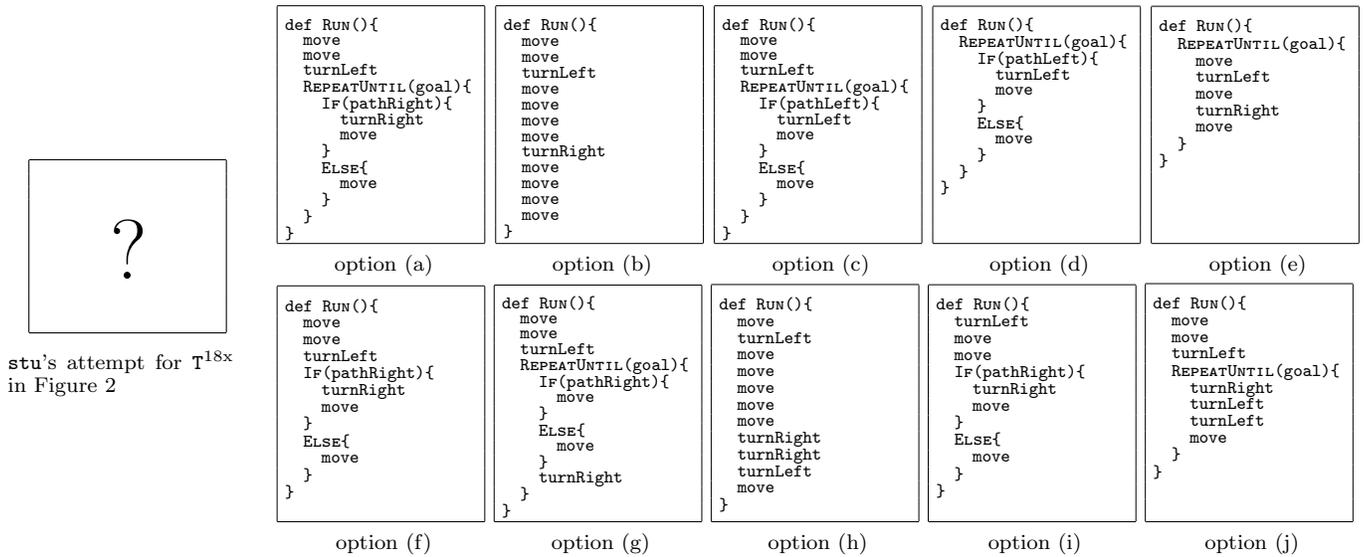

\centering
   \begin{minipage}{0.17\textwidth}
		\begin{subfigure}[b]{1\textwidth}
			\centering
			{
				\begin{boxcode}{3.0cm}{0.80}{0.58}
					\\
					\\
					\\
					\vspace{2mm}			
					\hspace{1.1cm} {\fontsize{40}{1}\selectfont ?}
					\\
					\\
					\vspace{4mm}
				\end{boxcode}
				\vspace{-3mm}
				\caption*{\student{}'s attempt for $\task^{\textnormal{18x}}$ in \figref{fig:intro.hoc18}
				}
				\label{fig:benchmark.hoc18.discriminative.unknown}
		    }
		    \end{subfigure}
   \end{minipage}
   \ \ 
   \begin{minipage}{0.81\textwidth}
			\begin{subfigure}[b]{.19\textwidth}
			\centering
			{
				\begin{boxcode}{3.4cm}{0.75}{0.72}
                    \textcode{def }\DSLRun\textcode{()}\textcode{\{}\\
                	\quad\DSLMove\\
                	\quad\DSLMove\\
                	\quad\DSLTurnLeft\\
                	\quad\DSLRepeatUntil\textcode{(\DSLBoolGoal)}\textcode{\{}\\
               	 	\quad\quad\DSLIf\textcode{(\DSLBoolPathRight)}\textcode{\{}\\
                	\quad\quad\quad\DSLTurnRight\\
                    \quad\quad\quad\DSLMove\\
                	\quad\quad\textcode{\}}\\
                	\quad\quad\DSLElse\textcode{\{}\\
                	\quad\quad\quad\DSLMove\\
               	 	\quad\quad\textcode{\}}\\
                	\quad\textcode{\}}\\
                	\textcode{\}}
				\end{boxcode}
				\vspace{-4mm}
				\caption*{\qquad option (a)} 
				\label{fig:benchmark.hoc18.discriminative.a}
		    }
		    \end{subfigure}
			\ 		   
			\begin{subfigure}[b]{.19\textwidth}
			\centering
			{

				\begin{boxcode}{3.4cm}{0.75}{0.76}
                      \textcode{def }\DSLRun\textcode{()}\textcode{\{}\\
                    \quad\DSLMove\\
                    \quad\DSLMove\\
                    \quad\DSLTurnLeft\\
                    \quad\DSLMove\\
                    \quad\DSLMove\\
                    \quad\DSLMove\\
                    \quad\DSLMove\\
                    \quad\DSLTurnRight\\
                    \quad\DSLMove\\
                    \quad\DSLMove\\
                    \quad\DSLMove\\
                    \quad\DSLMove\\
                    \textcode{\}}
                    \vspace{0.2mm}
				\end{boxcode}
				\vspace{-4mm}
				\caption*{\qquad option (b)}
				\label{fig:benchmark.hoc18.discriminative.b}
		    }
		    \end{subfigure}
			\ 			
			\begin{subfigure}[b]{.19\textwidth}
			\centering
			{
				\begin{boxcode}{3.4cm}{0.75}{0.72}
                    \textcode{def }\DSLRun\textcode{()}\textcode{\{}\\
                	\quad\DSLMove\\
                	\quad\DSLMove\\
                	\quad\DSLTurnLeft\\
                	\quad\DSLRepeatUntil\textcode{(\DSLBoolGoal)}\textcode{\{}\\
                	\quad\quad\DSLIf\textcode{(\DSLBoolPathLeft)}\textcode{\{}\\
                	\quad\quad\quad\DSLTurnLeft\\
                    \quad\quad\quad\DSLMove\\
                	\quad\quad\textcode{\}}\\
                	\quad\quad\DSLElse\textcode{\{}\\
                	\quad\quad\quad\DSLMove\\
               	 	\quad\quad\textcode{\}}\\
                	\quad\textcode{\}}\\
                	\textcode{\}}
				\end{boxcode}
				\vspace{-4mm}
				\caption*{\qquad option (c)}
				\label{fig:benchmark.hoc18.discriminative.c}
		    }
		    \end{subfigure}
			\ 	
			\begin{subfigure}[b]{.19\textwidth}
			\centering
			{
				\begin{boxcode}{3.4cm}{0.75}{0.72}
                    \textcode{def }\DSLRun\textcode{()}\textcode{\{}\\
                \quad\DSLRepeatUntil\textcode{(\DSLBoolGoal)}\textcode{\{}\\
                \quad\quad\DSLIf\textcode{(\DSLBoolPathLeft)}\textcode{\{}\\
                \quad\quad\quad\DSLTurnLeft\\
                \quad\quad\quad\DSLMove\\
                \quad\quad\textcode{\}}\\
                \quad\quad\DSLElse\textcode{\{}\\
                \quad\quad\quad\DSLMove\\
                \quad\quad\textcode{\}}\\
                \quad\textcode{\}}\\
                \textcode{\}}\\                    
                \\
                \\
				\end{boxcode}
				\vspace{-4mm}
				\caption*{\qquad option (d)}
				\label{fig:benchmark.hoc18.discriminative.d}
		    }
		    \end{subfigure}
			\ 					
			\begin{subfigure}[b]{.19\textwidth}
			\centering
			{
	    	\begin{boxcode}{3.4cm}{0.75}{0.80}
            \textcode{def }\DSLRun\textcode{()}\textcode{\{}\\
                \quad\DSLRepeatUntil\textcode{(\DSLBoolGoal)}\textcode{\{}\\
                \quad\quad\DSLMove\\
                \quad\quad\DSLTurnLeft\\
                \quad\quad\DSLMove\\
                \quad\quad\DSLTurnRight\\                
                \quad\quad\DSLMove\\
                \quad\textcode{\}}\\
                \textcode{\}}\\                  
				\vspace{9.7mm}
				\end{boxcode}
		        \vspace{-4mm}
				\caption*{\qquad option (e)}  
				\label{fig:benchmark.hoc18.discriminative.e}
		    }
		    \end{subfigure}
			\ 								
			\begin{subfigure}[b]{.19\textwidth}
			\centering
			{
				\begin{boxcode}{3.4cm}{0.75}{0.80}
                    \textcode{def }\DSLRun\textcode{()}\textcode{\{}\\
                \quad\DSLMove\\
                \quad\DSLMove\\
                \quad\DSLTurnLeft\\
                \quad\DSLIf\textcode{(\DSLBoolPathRight)}\textcode{\{}\\
                \quad\quad\DSLTurnRight\\
                \quad\quad\DSLMove\\
                \quad\textcode{\}}\\
                \quad\DSLElse\textcode{\{}\\
                \quad\quad\DSLMove\\
                \quad\textcode{\}}\\
                \textcode{\}}
				\vspace{3.3mm}
				\end{boxcode}
				\vspace{-4mm}
				\caption*{\qquad option (f)}
				\label{fig:benchmark.hoc18.discriminative.f}
		    }
		    \end{subfigure}
			\ 	
			\begin{subfigure}[b]{.19\textwidth}
			\centering
			{
		    	\begin{boxcode}{3.4cm}{0.75}{0.71}
                        \textcode{def }\DSLRun\textcode{()}\textcode{\{}\\
                    \quad\DSLMove\\
                    \quad\DSLMove\\
                    \quad\DSLTurnLeft\\
                     \quad\DSLRepeatUntil\textcode{(\DSLBoolGoal)}\textcode{\{}\\
                    \quad\quad\DSLIf\textcode{(\DSLBoolPathRight)}\textcode{\{}\\
                    \quad\quad\quad\DSLMove\\
                    \quad\quad\textcode{\}}\\
                    \quad\quad\DSLElse\textcode{\{}\\
                    \quad\quad\quad\DSLMove\\
                    \quad\quad\textcode{\}}\\
                    \quad\quad\DSLTurnRight\\
                    \quad\textcode{\}}\\
                    \textcode{\}}     
					\vspace{0.1mm}
				\end{boxcode}
                \vspace{-4mm}
				\caption*{\qquad option (g)}
				\label{fig:benchmark.hoc18.discriminative.g}
		    }
		    \end{subfigure}
			\ 				
			\begin{subfigure}[b]{.19\textwidth}
			\centering
			{
				\begin{boxcode}{3.4cm}{0.75}{0.80}
                    \textcode{def }\DSLRun\textcode{()}\textcode{\{}\\
                	\quad\DSLMove\\
                	\quad\DSLTurnLeft\\
                	\quad\DSLMove\\
                	\quad\DSLMove\\
                	\quad\DSLMove\\
                	\quad\DSLMove\\
                	\quad\DSLMove\\
                	\quad\DSLTurnRight\\
                	\quad\DSLTurnRight\\
                	\quad\DSLTurnLeft\\
                	\quad\DSLMove\\
                	\textcode{\}}
					\vspace{1.1mm}
				\end{boxcode}
				\vspace{-4mm}
				\caption*{\qquad option (h)}
				\label{fig:benchmark.hoc18.discriminative.h}
		    }
		    \end{subfigure}
			\ 
			\begin{subfigure}[b]{.19\textwidth}
			\centering
			{
				\begin{boxcode}{3.4cm}{0.75}{0.80}
                        \textcode{def }\DSLRun\textcode{()}\textcode{\{}\\
                        \quad\DSLTurnLeft\\
                        \quad\DSLMove\\
                        \quad\DSLMove\\
                        \quad\DSLIf\textcode{(\DSLBoolPathRight)}\textcode{\{}\\
                        \quad\quad\DSLTurnRight\\
                        \quad\quad\DSLMove\\
                        \quad\textcode{\}}\\
                        \quad\DSLElse\textcode{\{}\\
                        \quad\quad\DSLMove\\
                        \quad\textcode{\}}\\
                        \textcode{\}}
                        \vspace{3.5mm}
				\end{boxcode}
				\vspace{-4mm}
				\caption*{\qquad option (i)}
				\label{fig:benchmark.hoc18.discriminative.i}
		    }
		    \end{subfigure}
			\	
			\begin{subfigure}[b]{.19\textwidth}
			\centering
			{
                \begin{boxcode}{3.4cm}{0.75}{0.80}
                    \textcode{def }\DSLRun\textcode{()}\textcode{\{}\\
                \quad\DSLMove\\
                \quad\DSLMove\\
                \quad\DSLTurnLeft\\
                \quad\DSLRepeatUntil\textcode{(\DSLBoolGoal)}\textcode{\{}\\
                \quad\quad\DSLTurnRight\\
                \quad\quad\DSLTurnLeft\\
                \quad\quad\DSLTurnLeft\\
                \quad\quad\DSLMove\\
                \quad\textcode{\}}\\
                \textcode{\}}\\                
                \\
                \vspace{1mm}
				\end{boxcode}
				\vspace{-4mm}
				\caption*{\qquad option (j)}
				\label{fig:benchmark.hoc18.discriminative.j}
		    }
		    \end{subfigure}
			\					
   \end{minipage}   
    \caption{Illustration of the generative and discriminative objectives in the \benchmark{} benchmark for the scenario shown in Figure~\ref{fig:intro.hoc18}. For the generative objective, the goal is to synthesize the student \student{}'s behavior on the target task $\task^{\textnormal{18x}}$, i.e., a program that \student{} would write if the system would assign $\task^{\textnormal{18x}}$ to the student. For the discriminative objective, the goal is to choose one of the ten codes, shown as options (a)--(j), that corresponds to the student \student{}'s attempt. For each scenario, ten options are created systematically as discussed in Section~\ref{sec:benchmark.measures}; in this illustration, option (a) corresponds to the solution code $\code^{*}_{\task^\textnormal{18x}}$ for the target task and option (e) corresponds to the student \student{}'s attempt as designed in the benchmark. 
    }
	%
	\label{fig:benchmark.hoc18.discriminative}
\end{figure*}


\section{Benchmark and Initial Results}\label{sec:benchmark}
In this section, we introduce our benchmark, \benchmark{}, and report initial results highlighting the gap in performance of simple baselines and human experts.

\subsection{\benchmarktitle: Data Curation}\label{sec:benchmark.data}
We begin by curating a synthetic dataset for the benchmark, designed to capture different scenarios of the three distinct phases mentioned in Section~\ref{sec:problem.objectives}. In particular, each scenario corresponds to a 4-tuple $(\task^{\text{ref}}, \code^{\text{\student}}_{\task^{\text{ref}}}, \task^{\text{tar}}, \code^{\text{\student}}_{\task^{\text{tar}}})$, where $\code^{\text{\student}}_{\task^{\text{ref}}}$ (observed by the system) and $\code^{\text{\student}}_{\task^{\text{tar}}}$ (to be synthesized by the system) correspond to a student \student's attempts.

\looseness-1\textbf{Reference and target tasks.} 
We select two reference tasks for this benchmark, namely $\task^{4}$ and $\task^{18}$, as illustrated in Figures~\ref{fig:overview.hoc4.reftask}~and~\ref{fig:overview.hoc18.reftask}. These tasks correspond to Maze\#$4$ and Maze\#$18$ in the \emph{Hour of Code: Maze Challenge}~\cite{hourofcode_maze}, and have been studied in a number of prior works~\cite{DBLP:conf/lats/PiechSHG15,DBLP:conf/edm/EfremovGS20,DBLP:conf/nips/AhmedCEFGRS20}, because of the availability of large-scale datasets of students' attempts for these two tasks.
For each reference task, we manually create three target tasks as shown in Figures~\ref{fig:benchmark.hoc4.targettasks}~and~\ref{fig:benchmark.hoc18.targettasks}; as discussed in the figure captions, these target tasks are similar to the corresponding reference task in a sense that the set of available block types is same and the nesting structure of programming constructs in solution codes is same.
%

\textbf{Types of students' behaviors and students' attempts.} For a given reference-target task pair $(\task^{\text{ref}}, \task^{\text{tar}})$, next we seek to simulate a student \student to create \student's attempts $\code^{\text{\student}}_{\task^{\text{ref}}}$ and $\code^{\text{\student}}_{\task^{\text{tar}}}$. We begin by identifying a set of salient students' behaviors and misconceptions for reference tasks $\task^{\text{4}}$ and $\task^{\text{18}}$ based on students' attempts observed in the real-world dataset of~\cite{DBLP:conf/lats/PiechSHG15}. In this benchmark, we select $6$ types of students' behaviors for each reference task---these types are highlighted in Figures~\ref{fig:benchmark.hoc4.students}~and~\ref{fig:benchmark.hoc18.students} for $\task^{\text{4}}$ and $\task^{\text{18}}$, respectively.\footnote{\looseness-1In real-world settings, the types of students' behaviors and their attempts have a much larger variability and complexities with a long-tail distribution; in future work, we plan to extend our benchmark to cover more scenarios, see Section~\ref{sec:conclusions}.}
For a given pair $(\task^{\text{ref}}, \task^{\text{tar}})$, we first simulate a student \student by associating this student to one of the $6$ types, and then manually create \student's attempts $\code^{\text{\student}}_{\task^{\text{ref}}}$ and $\code^{\text{\student}}_{\task^{\text{tar}}}$. For a given scenario $(\task^{\text{ref}}, \code^{\text{\student}}_{\task^{\text{ref}}}, \task^{\text{tar}}, \code^{\text{\student}}_{\task^{\text{tar}}})$, the attempt $\code^{\text{\student}}_{\task^{\text{tar}}}$ is not observed and serves as a \emph{ground truth} in our benchmark for evaluation purposes; in the following, we interchangeably write a scenario as $(\task^{\text{ref}}, \code^{\text{\student}}_{\task^{\text{ref}}}, \task^{\text{tar}}, \text{\qmark})$.
%

\textbf{Total scenarios.} We create $72$ scenarios $(\task^{\text{ref}}, \code^{\text{\student}}_{\task^{\text{ref}}}, \task^{\text{tar}}, \code^{\text{\student}}_{\task^{\text{tar}}})$ in the benchmark corresponding to (i) $2$ reference tasks, (ii) $3$ target tasks per reference task, (iii) $6$ types of students' behaviors per reference task, and (iv) $2$ students per type. This, in turn, leads to a total of $72$ ($=2\times3\times6\times2$) unique scenarios.

\subsection{\benchmarktitle: Performance Measures} \label{sec:benchmark.measures}
We introduce two performance measures to capture our synthesis objective. Our first measure, namely \emph{generative performance}, is to directly capture the quality of fine-level synthesis of the student \student's attempt---this measure requires a human-in-the-loop evaluation. To further automate the evaluation process, we then introduce a second performance measure, namely \emph{discriminative performance}.
%

\textbf{Generative performance.}
As a generative performance measure, we introduce a $4$-point \textit{Likert scale} to evaluate the quality of synthesizing \student's attempt $\code^{\text{\student}}_{\task^{\text{tar}}}$ for a scenario $(\task^{\text{ref}}, \code^{\text{\student}}_{\task^{\text{ref}}}, \task^{\text{tar}}, \text{\qmark})$. The scale is designed to assign scores based on two factors: (a) whether the elements of the student's behavior observed in $\code^{\text{\student}}_{\task^{\textnormal{ref}}}$ are present, (b) whether the elements of the target task $\task^{\textnormal{tar}}$ (e.g., parts of its solution) are present. More concretely, the scores are assigned as follows (with higher scores being better): (i) Score $1$ means the technique does not have synthesis capability; (ii) Score $2$ means the synthesis fails to capture the elements of $\code^{\text{\student}}_{\task^{\textnormal{ref}}}$ and  $\task^{\textnormal{tar}}$; (iii) Score $3$ means the synthesis captures the elements only of $\code^{\text{\student}}_{\task^{\textnormal{ref}}}$ or of $\task^{\textnormal{tar}}$, but not both; (iv) Score $4$ means the synthesis captures the elements of both $\code^{\text{\student}}_{\task^{\textnormal{ref}}}$ and $\task^{\textnormal{tar}}$.
%

\textbf{Discriminative performance.} 
As the generative performance requires human-in-the-loop evaluation, we also introduce a disciminative performance measure based on the prediction accuracy of choosing the student attempt from a set.
More concretely, given a scenario $(\task^{\text{ref}}, \code^{\text{\student}}_{\task^{\text{ref}}}, \task^{\text{tar}}, \text{\qmark})$, the discriminative objective is to choose $\code^{\text{\student}}_{\task^{\textnormal{tar}}}$ from ten candidate codes; see Figure~\ref{fig:benchmark.hoc18.discriminative}. These ten options are created automatically in a systematic way and include the following: (a) the \emph{ground-truth} $\code^{\text{\student}}_{\task^{\textnormal{tar}}}$ from the benchmark, (b) the solution code  $\code^{\star}_{\task^{\textnormal{tar}}}$, (c) five codes $\code^{\text{\student}'}_{\task^{\textnormal{tar}}}$ from the benchmark associated with other students \ensuremath{\text{\student}'} whose behavior type is different from \student, and (iv) three \emph{randomly} constructed codes obtained by editing the solution code $\code^*_{\task^{\textnormal{tar}}}$.
%

\begin{table}[t!]
 \centering
 	\scalebox{0.76}{
 	\setlength\tabcolsep{2pt}
 	\renewcommand{\arraystretch}{1.0}
 	\begin{tabular}{r||p{2.15cm}p{2.15cm}||p{2.15cm}p{2.15cm}}
 		\toprule 
         \textbf{Method} & \multicolumn{2}{c||}{\textbf{Generative Performance}} & \multicolumn{2}{c}{\textbf{Discriminative Performance}}\\
          &   \multicolumn{1}{c}{Reference task} &  \multicolumn{1}{c||}{Reference task} &   \multicolumn{1}{c}{Reference task} &  \multicolumn{1}{c}{Reference task}\\        
          &  \multicolumn{1}{c}{$\task^{\textnormal{4}}$} & \multicolumn{1}{c||}{$\task^{\textnormal{18}}$} &  \multicolumn{1}{c}{$\task^{\textnormal{4}}$} & \multicolumn{1}{c}{$\task^{\textnormal{18}}$}\\        
 		\midrule
 		\algoRandom\textsuperscript{\phantom{1}}         & \multicolumn{1}{c}{$1.00$} & \multicolumn{1}{c||}{$1.00$} & \multicolumn{1}{c}{$10.15$} & \multicolumn{1}{c}{$10.10$}\\
 		\algoEditDist\textsuperscript{\phantom{1}}  & \multicolumn{1}{c}{$1.00$} & \multicolumn{1}{c||}{$1.00$} & \multicolumn{1}{c}{$30.83$} & \multicolumn{1}{c}{$47.06$}\\
 		\algoEmbDist\textsuperscript{\phantom{1}}     & \multicolumn{1}{c}{$1.00$} & \multicolumn{1}{c||}{$1.00$} & \multicolumn{1}{c}{$42.94$} & \multicolumn{1}{c}{$47.11$}\\
 		\midrule
        \algoTutor\textsuperscript{\phantom{1}}  & \multicolumn{1}{c}{$3.85$} & \multicolumn{1}{c||}{$3.91$} & \multicolumn{1}{c}{$89.81$} & \multicolumn{1}{c}{$85.19$}\\

        \color{TutorColour}
        \algoTutor\textsuperscript{1}  & \multicolumn{1}{c}{
        \color{TutorColour} $3.89$  } & \multicolumn{1}{c||}{
        \color{TutorColour} $3.94$} & \multicolumn{1}{c}{
        \color{TutorColour} $91.67$} & \multicolumn{1}{c}{
        \color{TutorColour} $83.33$}\\

        \color{TutorColour}
        \algoTutor\textsuperscript{2}  & \multicolumn{1}{c}{
        \color{TutorColour} $3.72$} & \multicolumn{1}{c||}{
        \color{TutorColour} $3.89$} & \multicolumn{1}{c}{
        \color{TutorColour} $91.67$} & \multicolumn{1}{c}{
        \color{TutorColour} $88.89$}\\

        \color{TutorColour}
        \algoTutor\textsuperscript{3}  & \multicolumn{1}{c}{
        \color{TutorColour} $3.94$} & \multicolumn{1}{c||}{
        \color{TutorColour} $3.89$} & \multicolumn{1}{c}{
        \color{TutorColour} $86.11$} & \multicolumn{1}{c}{
        \color{TutorColour} $83.33$}\\

 		\bottomrule
 	\end{tabular}
 	}
 	\caption{This table shows initial results on \benchmark{} in terms of the generative and discriminative performance measures. The values are in the range $[1.0, 4.0]$ for generative performance and in the range $[0.0, 100.0]$ for discriminative performance---higher values being better. Human experts (\algoTutor) can achieve high performance on both the measures, whereas simple baselines perform poorly. The numbers reported for \algoTutor{} are computed by averaging across three separate human experts (\algoTutor\textsuperscript{1}, \algoTutor\textsuperscript{2}, and \algoTutor\textsuperscript{3}). See Section~\ref{sec:benchmark.initialresults} for details.
 	}
     \label{table:benchmark.initialresults}
 \end{table}
 
%

\subsection{Initial Results}\label{sec:benchmark.initialresults}
As a starting point, we design a few simple baselines and compare their performance with that of human experts. %

%
\textbf{Simple baselines.} 
The simple baselines that we develop here are meant for the discriminative-only objective; they do not have  synthesis capability. Our first baseline \algoRandom{} simply chooses a code from the $10$ options at random.  Our next two baselines, \algoEditDist{} and \algoEmbDist{}, are defined through a distance function $\text{D}_{\task^{\textnormal{ref}}}(\code, \code')$ that quantifies a notion of distance between any two codes $\code, \code'$  for a fixed reference task. For a scenario $(\task^{\text{ref}}, \code^{\text{\student}}_{\task^{\text{ref}}}, \task^{\text{tar}}, \text{\qmark})$ and ten option codes, these baselines select the code  \code{} that minimizes $\text{D}_{\task^{\textnormal{ref}}}(\code, \code^{\text{\student}}_{\task^{\text{ref}}})$. \algoEditDist{} uses a tree-edit distance between \emph{Abstract Syntax Trees} as the distance function, denoted as $\text{D}^{\textnormal{edit}}_{\task^{\textnormal{ref}}}$. \algoEmbDist{} extends \algoEditDist{} by considering a distance function that combines $\text{D}^{\textnormal{edit}}_{\task^{\textnormal{ref}}}$ and a code-embedding based distance function $\text{D}^{\textnormal{emb}}_{\task^{\textnormal{ref}}}$; in this paper, we trained code embeddings with the methodology of \cite{DBLP:conf/edm/EfremovGS20} using a real-world dataset of student attempts on $\task^{\textnormal{ref}}$. \algoEmbDist{} then uses a distance function as a convex combination $\big(\alpha \cdot \text{D}^{\textnormal{edit}}_{\task^{\textnormal{ref}}}(\code, \code') + (1-\alpha)\cdot  \text{D}^{\textnormal{emb}}_{\task^{\textnormal{ref}}}(\code, \code')\big)$ where $\alpha$ is optimized for each reference task separately. For measuring the discriminative performance, we randomly sample a scenario, create ten options, and measure the predictive accuracy of the technique---the details of this experimental evaluation are provided in Section~\ref{sec:experiments:dicriminative}. %

\looseness-1\textbf{Human experts.} Next, we evaluate the performance of human experts on the benchmark \benchmark, and refer to this evaluation technique as \algoTutor{}. These evaluations are done through a web platform where an expert would provide a generative or discriminative response to a given scenario $(\task^{\text{ref}}, \code^{\text{\student}}_{\task^{\text{ref}}}, \task^{\text{tar}}, \text{\qmark})$. In our work, \algoTutor{} involved participation of three independent experts for the evaluation; these experts have had experience in block-based programming and tutoring. We first carried out generative performance evaluations where an expert had to write the student attempt code; afterwards, we carried out discriminative performance evaluations where an expert would choose one of the options. In total, each expert participated in $36$ generative evaluations ($18$ per reference task) and $72$ discriminative evaluations ($36$ per reference task). Results in Table~\ref{table:benchmark.initialresults} highlight the huge performance gap between the human experts and simple baselines; further details are provided in Section~\ref{sec:experiments}.

\section{Neural Synthesizer \algoContinualtitle{}}
\label{sec:continual}
\begin{figure*}[t!]
\captionsetup{singlelinecheck = false, format= hang, justification=raggedright}
\centering
	\includegraphics[width=1\textwidth]{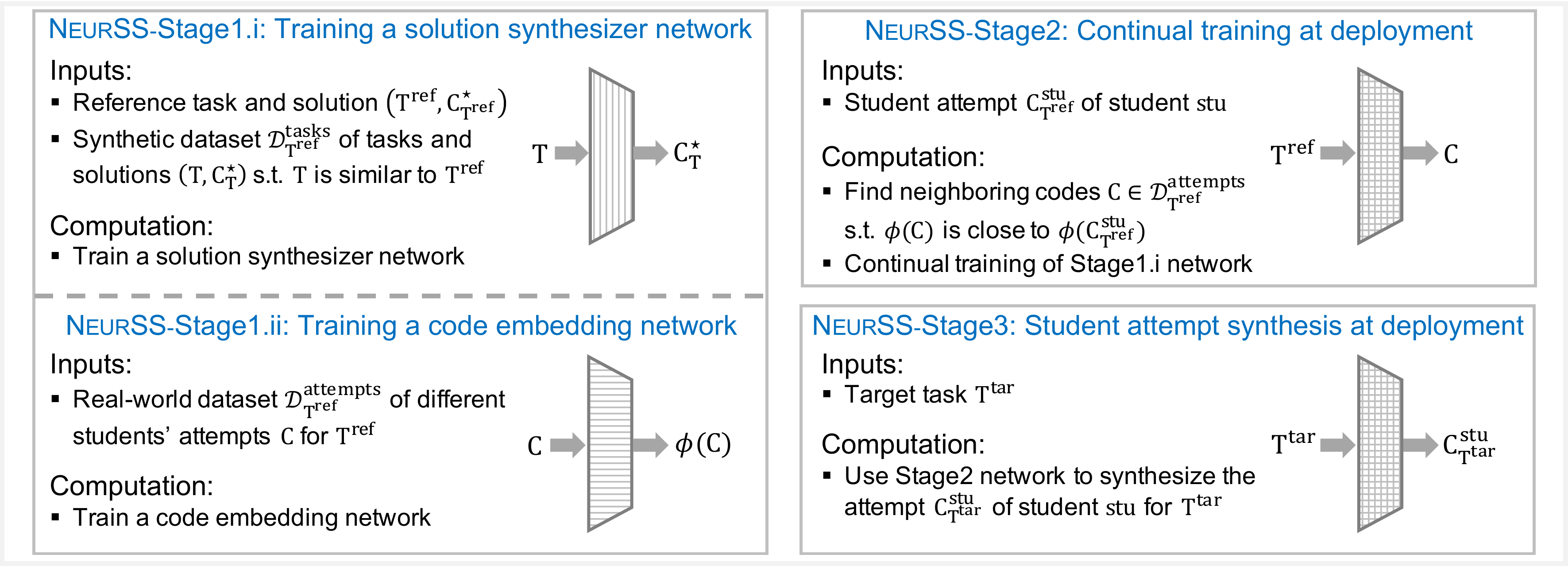}	
	\caption{Illustration of the three different stages in \algoContinual{}, our technique based on neural synthesis; details in Section~\ref{sec:continual}.
	}
	\vspace{2mm}
	\label{fig:continual.pipeline}
\end{figure*}

Our first technique, \algoContinual{} (\emph{Neural Program Synthesis for \benchmark}), is inspired by recent advances in neural program synthesis~\cite{DBLP:conf/iclr/BunelHDSK18,DBLP:conf/iclr/ChenLS19}. In our work, we use the neural architecture proposed in~\cite{DBLP:conf/iclr/BunelHDSK18}---at a high-level, the neural synthesizer model takes as input a visual task $\task$, and then sequentially synthesizes a code $\code$ by using programming tokens in  $\task_{\textnormal{store}}$. However, our goal is not simply to synthesize a solution code, instead, we want to synthesize attempts of a given student that the system is interacting with at real-time/deployment. To achieve this goal, \algoContinual{} operates in three stages as illustrated in Figure~\ref{fig:continual.pipeline}. Each stage is in line with a phase of our objective described in Section~\ref{sec:problem.objectives}. 
At a high-level, the three stages of \algoContinual{} are as follows: (i) In Stage1, we are given a reference task and its solution $(\task^{\textnormal{ref}}, \code^{\star}_{\task^{\textnormal{ref}}})$, and train a neural synthesizer model that can synthesize solutions for any task similar to  $\task^{\textnormal{ref}}$; (ii) In Stage2, the system observes the student \student's attempt $\code^{\text{\student}}_{\task^{\textnormal{ref}}}$ and initiates \emph{continual training} of the neural synthesizer model from Stage1 in real-time; (iii) In Stage3, the system considers a target task $\task^{\textnormal{tar}}$ and uses the model from Stage2 to synthesize $\code^{\text{\student}}_{\task^{\textnormal{tar}}}$.
In the following paragraphs, we provide an overview of the key ideas and high-level implementation details for each stage.
%

%
%

\textbf{\algoContinualtextbf{}-Stage1.i.} Given a reference task and its solution $(\task^{\textnormal{ref}}, \code^{\star}_{\task^{\textnormal{ref}}})$, the goal of this stage is to train a neural synthesizer model that can synthesize solutions for any task similar to  $\task^{\textnormal{ref}}$.  In this stage, we use a synthetic dataset $\mathcal{D}^{\text{tasks}}_{\task^{\text{ref}}}$ comprising of task-solution pairs $(\task, \code^{\star}_{\task})$; the notion of similarity here means that $\task_\textnormal{store}$ is the same as $\task^{\textnormal{ref}}_\textnormal{store}$ and the nesting structure of programming constructs in $\code^{\star}_{\task}$ is the same as in $\code^{\star}_{\task^{\textnormal{ref}}}$. To train this synthesizer, we leverage recent advances in neural program synthesis~\cite{DBLP:conf/iclr/BunelHDSK18, DBLP:conf/iclr/ChenLS19}; in particular, we use the encoder-decoder architecture and imitation learning procedure from \cite{DBLP:conf/iclr/BunelHDSK18}. The model we use in our experiments has  deep-CNN layers for extracting task features and an LSTM for sequentially generating programming tokens. The input to the synthesizer is a one-hot task representation of the visual grid denoting different elements of the grid (e.g., ``goal'', ``walls'', and position/orientation of the ``avatar''), as well as the programming tokens synthesized by the model so far. To generate the synthetic dataset $\mathcal{D}^{\text{tasks}}_{\task^{\text{ref}}}$, we use the  task generation procedure from \cite{DBLP:conf/nips/AhmedCEFGRS20}. For the reference task $\task^{\textnormal{4}}$, we generated $\mathcal{D}^{\text{tasks}}_{\task^{\text{4}}}$ of size $50,000$; for the reference task $\task^{\textnormal{18}}$, we generated $\mathcal{D}^{\text{tasks}}_{\task^{\text{18}}}$ of size $200,000$.

\textbf{\algoContinualtextbf{}-Stage1.ii.} Given a reference task $\task^{\textnormal{ref}}$, the goal of this stage is to train a code embedding network that maps an input code $\code$ to a feature vector $\phi(\code)$. This code embedding space will be useful later in \algoContinualtextbf{}-Stage2 when we observe the student \student's attempt. For each $\task^{\textnormal{ref}}$, we use a real-world dataset of students' attempts $\mathcal{D}^{\text{attempts}}_{\task^{\text{ref}}}$ on $\task^{\textnormal{ref}}$ to train this embedding network using the methodology of \cite{DBLP:conf/edm/EfremovGS20}. To train this embedding network, we construct a set with triplets $(\code, \code', \text{D}^{\textnormal{edit}}_{\task^{\textnormal{ref}}}(\code,\code'))$ where $\code, \code' \in \mathcal{D}^{\text{attempts}}_{\task^{\text{ref}}}$ and $\text{D}^{\textnormal{edit}}_{\task^{\textnormal{ref}}}$ computes the tree-edit distance between \emph{Abstract Syntax Trees} of two codes (see Section~\ref{sec:benchmark.initialresults}). The embedding network is trained so the embedding space preserves given distances, i.e., $||\phi(\code) - \phi(\code')|| \approx \text{D}_{\task^{\textnormal{ref}}}^{\textnormal{edit}}(\code,\code')$ for a triplet. Following the setup in \cite{DBLP:conf/edm/EfremovGS20}, we use a bidirectional LSTM architecture for the network and use $\mathbb{R}^{80}$ embedding space. 
%

\textbf{\algoContinualtextbf{}-Stage2.} In this stage, the system observes the student \student's attempt $\code^{\text{\student}}_{\task^{\textnormal{ref}}}$ and initiates continual training of the neural synthesizer model from Stage1.i in real-time. More concretely, we fine-tune the pre-trained synthesizer model from Stage 1.i with the goal of transferring the student \student's behavior from  the reference task $\task^{\text{ref}}$ to any target task $\task^{\textnormal{tar}}$. Here, we make use of the embedding network from Stage1.ii that enables us to find neighboring codes $\code \in \mathcal{D}^{\text{attempts}}_{\task^{\text{ref}}}$ such that $\phi(\code)$ is close to $\phi(\code^{\text{\student}}_{\task^{\textnormal{ref}}})$. More formally, the set of neighbors is given by $\{ \code \in  \mathcal{D}^{\text{attempts}}_{\task^{\text{ref}}} : ||\phi(\code^{\text{\student}}_{\task^{\textnormal{ref}}}) - \phi(\code)||_2 \leq r\}$ where the threshold $r$ is a hyperparameter.
Next, we use these neighboring codes to create a small dataset for continual training: this dataset comprises of the task-code pairs $(\code, \task^{\text{ref}})$ where $\code$ is a neighboring code for $\code^{\text{\student}}_{\task^{\textnormal{ref}}}$ and $\task^{\text{ref}}$ is the reference task. There are two crucial ideas behind the design of this stage. First, we do this continual training using a set of neighboring codes w.r.t. $\code^{\text{\student}}_{\task^{\textnormal{ref}}}$ instead of just using $\code^{\text{\student}}_{\task^{\textnormal{ref}}}$---this is important to avoid overfitting during the process. Second, during this continual training, we train for a small number of epochs (a hyperparameter), and only fine-tune the decoder by freezing the encoder---this is important so that the network obtained after continual training still maintains its synthesis capability. The hyperparameters in this stage (threshold $r$, the number of epochs and learning rate) are obtained through cross-validation in our experiments (see Section~\ref{sec.experiments.discriminative})

\textbf{\algoContinualtextbf{}-Stage3.} In this stage, the system observes $\task^{\textnormal{tar}}$ and uses the model from Stage2 to synthesize $\code^{\text{\student}}_{\task^{\textnormal{tar}}}$. More concretely, we provide $\task^{\textnormal{tar}}$ as an input to the Stage2 model and then synthesize a small set of codes as outputs using a \emph{beam search} procedure proposed in \cite{DBLP:conf/iclr/BunelHDSK18}. This procedure allows us to output codes that have high likelihood or probability of synthesis with the model. In our experiments, we use a \emph{beam size} of $64$; Figures~\ref{fig:experiments.hoc4.generative.continual}~and~\ref{fig:experiments.hoc18.generative.continual} illustrate Top-$3$ synthesized codes for different scenarios obtained through this procedure. The Top-$1$ code is then used for generative performance evaluation. For the discriminative performance evaluation, we are given a set of option codes; here we use the model of Stage2 to compute the likelihood of provided options and then select one with the highest probability.

\section{Symbolic Synthesizer \algoPCFGtitle{}}
\label{sec:pcfg}

\begin{figure}[t!]
\centering
	\includegraphics[width=1\linewidth]{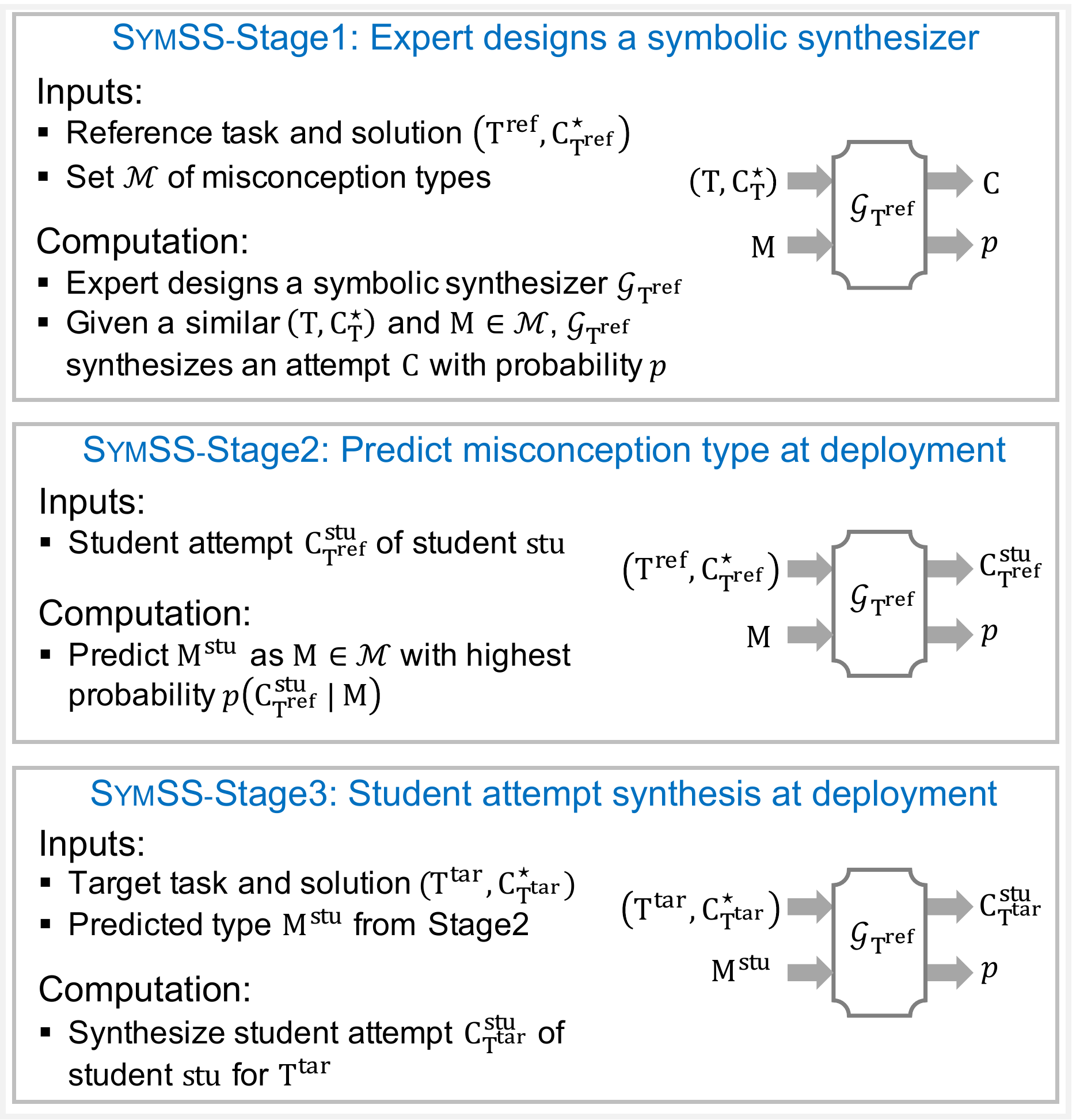}
    \caption{Illustration of the three different stages in \algoPCFG{}, our technique based on symbolic synthesis; see Section~\ref{sec:pcfg}.
	}
	\label{fig:pcfg.pipeline}
\end{figure}

In the previous section, we introduced \algoContinual{} inspired by neural program synthesis. \algoContinual{} additionally has synthesis capability in comparison to the simple baselines introduced earlier; yet, there is a substantial gap in the performance of \algoContinual{} and human experts (i.e., \algoTutor{}). 
%
An important question that we seek to resolve is how much of this performance gap can be reduced by leveraging domain knowledge such as  how students with different behaviors (misconceptions) write codes. To this end, we introduce our second technique, \algoPCFG{} (\emph{Symbolic Program Synthesis for \benchmark}), inspired by recent advances in using symbolic methods for program synthesis~\cite{lake2015human,DBLP:conf/aaai/WuMGP19,DBLP:conf/edm/MalikWVSCMGP21}. Similar in spirit to \algoContinual{}, \algoPCFG{} operates in three stages as illustrated in Figure~\ref{fig:pcfg.pipeline}. Each stage is in line with a phase of our objective described in Section~\ref{sec:problem.objectives}. At a high-level, the three stages of \algoPCFG{} are as follows: (i) In Stage1, we are given $(\task^{\textnormal{ref}}, \code^{\star}_{\task^{\textnormal{ref}}})$, and design a symbolic synthesizer model using \emph{Probabilistic Context Free Grammars} (PCFG)s to encode how students of different behavior types $\mathcal{M}$ write codes for any task similar to $\task^{\textnormal{ref}}$~\cite{Chomsky1959OnCF,martin1991introduction,DBLP:conf/aaai/WuMGP19}; (ii) In Stage2, the system observes the student \student's attempt $\code^{\text{\student}}_{\task^{\textnormal{ref}}}$ and makes a prediction about the  behavior type $\text{M}^{\text{\student}} \in \mathcal{M}$; (iii) In Stage3, the system considers a target task $\task^{\textnormal{tar}}$ and uses the model from Stage1 to synthesize $\code^{\text{\student}}_{\task^{\textnormal{tar}}}$ based on the inferred $\text{M}^{\text{\student}}$. 
In the following paragraphs, we provide an overview of the key ideas and high-level implementation details for each stage.

\begin{figure}[t!]
    \centering
	\begin{center}
	\scalebox{0.94}{
		\setlength\tabcolsep{4pt}
		\renewcommand{\arraystretch}{0.6}
    	\begin{tabular}{|p{0.15\linewidth}p{0.3\linewidth}p{0.15\linewidth}p{0.3\linewidth}|}  
 			\hline
			\multicolumn{1}{|p{0.15\linewidth}}{\GStart := } &  
			     \multicolumn{3}{p{0.45\linewidth}|}{$\xrightarrow{p_1}$ \GRight~\GMove~\GLeft~\GMove~\GMove~\GMove~\GLeft~\GMove}\\
			    & \multicolumn{3}{p{0.45\linewidth}|}{$\xrightarrow{p_2}$ \GRight~\GMove~\GLeft~\GMove~\GMove~\GMove~\GMove}\\
                & \multicolumn{3}{p{0.45\linewidth}|}{$\xrightarrow{p_2}$ \GRight~\GMove~\GMove~\GMove~\GMove~\GLeft~\GMove}\\
                & \multicolumn{3}{p{0.45\linewidth}|}{$\xrightarrow{p_2}$ \GMove~\GLeft~\GMove~\GMove~\GMove~\GLeft~\GMove}\\
                & \multicolumn{3}{p{0.45\linewidth}|}{$\xrightarrow{p_3}$ \GRight~\GMove~\GMove~\GMove~\GMove~\GMove} \\
                & \multicolumn{3}{p{0.45\linewidth}|}{$\xrightarrow{p_3}$  \GMove~\GLeft~\GMove~\GMove~\GMove~\GMove} \\
                & \multicolumn{3}{p{0.45\linewidth}|}{$\xrightarrow{p_3}$  \GMove~\GMove~\GMove~\GMove~\GLeft~\GMove } \\
                & \multicolumn{3}{p{0.45\linewidth}|}{$\xrightarrow{p_4}$ \GMove~\GMove~\GMove~\GMove~\GMove}\\

                & \multicolumn{3}{p{0.35\linewidth}|}{}\\

                \multicolumn{4}{|p{1\linewidth}|}{\textcolor{PCFGColour}{The solution code $\code^{\star}_{\task^{\textnormal{4x}}}$ for $\task^{\textnormal{4x}}$ is \{\DSLRun\{\DSLTurnRight; \DSLMove; \DSLTurnLeft; \DSLMove; \DSLMove; \DSLMove; \DSLTurnLeft; \DSLMove{}\}\}. These rules for \GStart are specific to the behavior type M$^{\text{\student}}$ that corresponds to forgetting to include some turns in the solution and are created automatically w.r.t. $\code^{\star}_{\task^{\textnormal{4x}}}$.}}\\ %
                \hline
                & \multicolumn{3}{p{0.35\linewidth}|}{}\\
                
                \GMove:=     &\hspace{-5mm}$\xrightarrow{p_5}$ \GRepMove~\GRepMove & \GRepMove :=   &$\xrightarrow{p_7}$ \GRepMove~\GRepMove\\
                            &\hspace{-5mm}$\xrightarrow{p_6}$ \DSLMove &&$\xrightarrow{p_7}$ \DSLMove\\
                            &\hspace{-5mm}$\xrightarrow{p_5}$ \DSLTurnLeft &&$\xrightarrow{p_8}$ \DSLTurnLeft\\
                            &\hspace{-5mm}$\xrightarrow{p_5}$ \DSLTurnRight &&$\xrightarrow{p_8}$ \DSLTurnRight\\
                & \multicolumn{3}{p{0.35\linewidth}|}{}\\

                \GLeft:=     &\hspace{-5mm}$\xrightarrow{p_5}$ \GRepLeft~\GRepLeft & \GRepLeft :=   &$\xrightarrow{p_7}$ \GRepLeft~\GRepLeft\\
                            &\hspace{-5mm}$\xrightarrow{p_5}$ \DSLMove &&$\xrightarrow{p_8}$ \DSLMove\\
                            &\hspace{-5mm}$\xrightarrow{p_6}$ \DSLTurnLeft &&$\xrightarrow{p_7}$ \DSLTurnLeft\\
                            &\hspace{-5mm}$\xrightarrow{p_5}$ \DSLTurnRight &&$\xrightarrow{p_8}$ \DSLTurnRight\\
                & \multicolumn{3}{p{0.35\linewidth}|}{}\\

                \GRight:=     &\hspace{-5mm}$\xrightarrow{p_5}$ \GRepRight~\GRepRight & \GRepRight :=   &$\xrightarrow{p_7}$ \GRepRight~\GRepRight\\
                            &\hspace{-5mm}$\xrightarrow{p_5}$ \DSLMove &&$\xrightarrow{p_8}$ \DSLMove\\
                            &\hspace{-5mm}$\xrightarrow{p_5}$ \DSLTurnLeft &&$\xrightarrow{p_8}$ \DSLTurnLeft\\
                            &\hspace{-5mm}$\xrightarrow{p_6}$ \DSLTurnRight &&$\xrightarrow{p_7}$ \DSLTurnRight\\

                & \multicolumn{3}{p{0.35\linewidth}|}{}\\

                \multicolumn{4}{|p{1\linewidth}|}{\textcolor{PCFGColour}{These rules are specific to the behavior type M$^{\text{\student}}$ and independent of the task and solution code $(\task, \code^{\star}_{\task})$.}}\\
			\hline
 		\end{tabular}
	}
\end{center}
\caption{
    PCFG corresponding to $\mathcal{G}_{\task^{\textnormal{4}}}(\task^{\textnormal{4x}}, \code^{\star}_{\task^{\textnormal{4x}}}, \textnormal{M}^{\textnormal{\student}})$---this PCFG is automatically created in \algoPCFGtextbf{}-Stage3 for the scenario in Figure~\ref{fig:intro.hoc4} and is used to synthesize student \student's attempt. We define the following symbols: (i) the start symbol \GStart; (ii) the non-terminal symbols  \GMove, \GLeft, \GRight, \GRepMove, \GRepLeft, \GRepRight; (iii) the terminal symbols \DSLMove, \DSLTurnLeft, \DSLTurnRight. For this scenario, the behavior type M$^{\text{\student}}$ corresponds to forgetting to include some turns in the solution. The production rules for \GStart are specific to the behavior type M$^{\text{\student}}$ and are automatically created w.r.t. the solution code $\code^{\star}_{\task^{\textnormal{4x}}}$, leading to the following rules: include all the turns ($p_1$), omit one turn ($p_2$), omit two turns ($p_3$), or omit all of the three turns ($p_4$). 
    We designed the probability of keeping a turn as $p_k = 1/3$ and omitting a turn as $p_o = 1 - p_k = 2/3$; this leads to the following probability values for \GStart's production rules: \{$p_1 = p_k^3,~p_2 = p_k^2 \cdot p_o,~p_3 = p_k \cdot p_o^2,~p_4 = p_o^3$\}.
    %
    The production rules for \GMove, \GLeft, \GRight, \GRepMove, \GRepLeft, \GRepRight are specific to the behavior type $\textnormal{M}^{\textnormal{\student}}$ and independent of the task and solution code $(\task, \code^{\star}_{\task})$ in $\mathcal{G}_{\task^{\textnormal{4}}}(\task, \code^{\star}_{\task}, \textnormal{M}^{\textnormal{\student}})$. The symbols \GRepMove, \GRepLeft, \GRepRight are introduced to allow variability in the students' attempts; intuitively \student{} may further diverge from the behavior type by (i) transforming tokens,  (ii) replicating tokens, or (iii) adding a sequence of tokens. For these rules, the probabilities values are defined as: \{$p_5 = 0.1,~p_6 = 0.7,~p_7 = 0.4,~p_8 = 0.1$\}.
     } 
\label{fig:pcfg.grammar}
\vspace{1mm}
\end{figure}

\textbf{\algoPCFGtextbf{}-Stage1 (High-level design).} For a given reference task and its solution $(\task^{\textnormal{ref}}, \code^{\star}_{\task^{\textnormal{ref}}})$, the goal of this stage is to create a symbolic program synthesizer that encodes domain knowledge. In this stage, we need access to the following: (i) a set of types of students' behaviors the system is expected to encounter at deployment, denoted by $\mathcal{M}$ in Figure~\ref{fig:pcfg.pipeline}; (ii) an expert with domain knowledge.\footnote{We note that \algoPCFG{} is the only technique that requires access to the types of students' behaviors; in our implementation and experiments, we considered $\mathcal{M}$ to be the same as the types of students' behaviors in \benchmark. In practice, there could potentially be a large number of types of behaviors, that manifest in students' attempts in an unknown way; hence, \algoPCFG{} in a real-world setting could perform worse than the performance reported on \benchmark. Also, we note that human experts in \algoTutor{} were not told about the types of students' behaviors in \benchmark.}
The expert then designs a symbolic program synthesizer  $\mathcal{G}_{\task^{\textnormal{ref}}}(\task,\code_{\task}^\star, \text{M})$ for the reference task $\task^{\textnormal{ref}}$ that operates as follows: (a) as first input, it takes a task-solution pair ($\task, \code^\star_{\task}$) where $\task$ is expected to be similar to $\task^{\textnormal{ref}}$; (b) as second input, it takes a type of student behavior $\text{M} \in \mathcal{M}$; (c) as outputs, it synthesizes a code $\code$ along with the probability $p$ of synthesis. This symbolic synthesizer $\mathcal{G}_{\task^{\text{ref}}}(\task,\code^\star_{\task},\text{M})$ is designed as a \emph{grammar creator module}: internally, it first automatically creates a specific \emph{grammar} corresponding to its input arguments and then generates codes based on this \emph{grammar}.

\begin{table*}[t!]
 \centering
 	\scalebox{0.82}{
 	\setlength\tabcolsep{2.8pt}
 	\renewcommand{\arraystretch}{1.0}
 	\begin{tabular}{r||p{2.15cm}p{2.15cm}||p{2.15cm}p{2.15cm}||p{2.55cm}p{2.55cm}p{1.35cm}p{1.35cm}p{1.35cm}}
 		\toprule 
         \textbf{Method} & \multicolumn{2}{c||}{\textbf{Generative Performance}} & \multicolumn{2}{c||}{\textbf{Discriminative Performance}} & \multicolumn{5}{c}{\textbf{Required Inputs and Domain Knowledge}} \\
          &   \multicolumn{1}{c}{Reference task} &  \multicolumn{1}{c||}{Reference task} &   \multicolumn{1}{c}{Reference task} &  \multicolumn{1}{c||}{Reference task} &  \multicolumn{1}{c}{Ref. task dataset:} &  \multicolumn{1}{c}{Ref. task dataset:} &  \multicolumn{1}{c}{Student} &  \multicolumn{1}{c}{Expert}  &  \multicolumn{1}{|c}{Expert} \\        
          &  \multicolumn{1}{c}{$\task^{\textnormal{4}}$} & \multicolumn{1}{c||}{$\task^{\textnormal{18}}$} &  \multicolumn{1}{c}{$\task^{\textnormal{4}}$} & \multicolumn{1}{c||}{$\task^{\textnormal{18}}$} & \multicolumn{1}{c}{student 
attempts} & \multicolumn{1}{c}{similar tasks} & \multicolumn{1}{c}{types} & \multicolumn{1}{c}{grammars} & \multicolumn{1}{|c}{evaluation} \\        
 		\midrule
 		\algoRandom        & \multicolumn{1}{c}{$1.00$} & \multicolumn{1}{c||}{$1.00$} & \multicolumn{1}{c}{$10.15 \pm 0.2$} & \multicolumn{1}{c||}{$10.10 \pm 0.2 $} & \multicolumn{1}{c}{-} & \multicolumn{1}{c}{-} & \multicolumn{1}{c}{-} & \multicolumn{1}{c}{-} & \multicolumn{1}{|c}{-} \\
 		\algoEditDist & \multicolumn{1}{c}{$1.00$} & \multicolumn{1}{c||}{$1.00$} & \multicolumn{1}{c}{$30.83 \pm 1.1$} & \multicolumn{1}{c||}{$47.06 \pm 0.3$} & \multicolumn{1}{c}{-} & \multicolumn{1}{c}{-} & \multicolumn{1}{c}{-} & \multicolumn{1}{c}{-} & \multicolumn{1}{|c}{-} \\
 		\algoEmbDist    & \multicolumn{1}{c}{$1.00$} & \multicolumn{1}{c||}{$1.00$} & \multicolumn{1}{c}{$42.94\pm 2.1$} & \multicolumn{1}{c||}{$47.11 \pm 0.8$} & \multicolumn{1}{c}{\xmark} & \multicolumn{1}{c}{-} & \multicolumn{1}{c}{-} & \multicolumn{1}{c}{-} & \multicolumn{1}{|c}{-} \\
 		\midrule
 		\algoContinual & \multicolumn{1}{c}{$3.28$} & \multicolumn{1}{c||}{$2.94$} & \multicolumn{1}{c}{$40.10\pm 0.7$} & \multicolumn{1}{c||}{$55.98\pm 1.5$}  & \multicolumn{1}{c}{\xmark} & \multicolumn{1}{c}{\xmark} & \multicolumn{1}{c}{-} & \multicolumn{1}{c}{-} & \multicolumn{1}{|c}{-} \\
 		\algoPCFG  & \multicolumn{1}{c}{$3.72$} & \multicolumn{1}{c||}{$3.83$} & \multicolumn{1}{c}{$87.17\pm 0.7$} & \multicolumn{1}{c||}{$67.83\pm 1.0 $}  & \multicolumn{1}{c}{-} & \multicolumn{1}{c}{-} & \multicolumn{1}{c}{\xmark} & \multicolumn{1}{c}{\xmark} & \multicolumn{1}{|c}{-}\\
 		\midrule
        \algoTutor  & \multicolumn{1}{c}{$3.85$} & \multicolumn{1}{c||}{$3.91$} & \multicolumn{1}{c}{$89.81\pm 1.9$} & \multicolumn{1}{c||}{$85.19\pm 1.9$}  & \multicolumn{1}{c}{-} & \multicolumn{1}{c}{-} & \multicolumn{1}{c}{-} & \multicolumn{1}{c}{-} & \multicolumn{1}{|c}{\xmark} \\
 		\bottomrule
 	\end{tabular}
 	}
 	\caption{This table expands on Table~\ref{table:benchmark.initialresults} and additionally provides results for \algoContinual{} and \algoPCFG{}. The columns under ``Required Inputs and Domain Knowledge'' highlight information used by different techniques (\xmark{} indicates the usage of the corresponding input/knowledge). \algoContinual{} and \algoPCFG{} significantly improve upon the simple baselines introduced in Section~\ref{sec:benchmark.initialresults}; yet, there is a gap in performance in comparison to that of human experts. See Section~\ref{sec:experiments} for details.}
     \label{table:benchmark.finalresults}
 \end{table*}

%

\textbf{\algoPCFGtextbf{}-Stage1 (PCFG).} 
Inspired by recent work on modeling students' misconceptions via \emph{Probabilistic Context Free Grammars} (PCFG)s \cite{DBLP:conf/aaai/WuMGP19}, we consider a PCFG family of grammars inside $\mathcal{G}_{\task^{\text{ref}}}$.\footnote{Context Free Grammars (CFG)s generate strings by applying a set of production rules where each symbol is expanded independently of its context~\cite{martin1991introduction}. These rules are defined through a \emph{start} symbol, \emph{non-terminal} symbols, and \emph{terminal} symbols. PCFGs additionally assign a probability to each production rule; see Figure~\ref{fig:pcfg.grammar} as an example.}
More concretely, given a reference task $\task^{\textnormal{ref}}$, a task-solution pair $(\task, \code^\star_{\task})$, and a type M, the expert has designed an automated function that creates a PCFG corresponding to $\mathcal{G}_{\task^{\textnormal{ref}}}(\task,\code_{\task}^\star, \text{M})$ which is then used to sample/synthesize codes. This PCFG is created automatically and the production rules are based on: the type M, the input solution code $\code_{\task}^\star$, and optionally features of $\task$. In our implementation, we designed two separate symbolic synthesizers $\mathcal{G}_{\task^{\text{4}}}$ and $\mathcal{G}_{\task^{\text{18}}}$ associated with two reference tasks. As a concrete example, consider the scenario in Figure~\ref{fig:intro.hoc4}: the PCFG created internally at \algoPCFG{}-Stage3 corresponds to $\mathcal{G}_{\task^{\textnormal{4}}}(\task^{\textnormal{4x}}, \code^{\star}_{\task^{\textnormal{4x}}}, \textnormal{M}^{\textnormal{\student}})$ and is illustrated in Figure~\ref{fig:pcfg.grammar}; details are provided in the caption and as comments within the figure.

\textbf{\algoPCFGtextbf{}-Stage2}.
In this stage, the system observes the student \student's attempt $\code^{\text{\student}}_{\task^{\textnormal{ref}}}$ and makes a prediction about the  behavior type $\text{M}^{\text{\student}} \in \mathcal{M}$. For each behavior type $\text{M} \in \mathcal{M}$ specified at Stage1, we use $\mathcal{G}_{\task^{\textnormal{ref}}}$ with arguments $(\task^{\text{ref}}, \code^{\star}_{\textnormal{ref}}, \text{M})$ to calculate the probability of synthesizing $\code^{\text{\student}}_{\task^{\text{ref}}}$ w.r.t. $\text{M}$, referred to as $p(\code^{\text{\student}}_{\task^{\text{ref}}}|\text{M})$. This is done by internally creating a corresponding PCFG for $\mathcal{G}_{\task^{\textnormal{ref}}}(\task^{\text{ref}}, \code^{\star}_{\textnormal{ref}}, \text{M})$. To predict M$^{\text{\student}}$, we pick the behavior type M with the highest probability. As an implementation detail, we construct PCFGs in a special form called the \emph{Chmosky Normal Form} (CNF) \cite{Chomsky1959OnCF,martin1991introduction} (though the PCFG illustrated in Figure~\ref{fig:pcfg.grammar} is not in CNF).
This form imposes constraints to the grammar rules that add extra difficulty in grammar creation, but enables the efficient calculation of $p(\code^{\text{\student}}_{\task^{\text{ref}}}|\text{M})$.

\looseness-1\textbf{\algoPCFGtextbf{}-Stage3.} In this stage, the system observes a target task $\task^{\textnormal{tar}}$ along with its solution $\code^{\star}_{\task^{\textnormal{tar}}}$. Based on the behavior type $\text{M}^{\text{\student}}$ inferred in Stage2, it uses $\mathcal{G}_{\task^{\textnormal{ref}}}$ with input arguments $(\task^{\textnormal{tar}}, \code_{\task^{\textnormal{tar}}}^\star,\text{M}^{\text{\student}})$ to synthesize $\code^{\text{\student}}_{\task^{\textnormal{tar}}}$. More concretely, we use $\mathcal{G}_{\task^{\textnormal{ref}}}(\task^{\textnormal{tar}}, \code_{\task^{\textnormal{tar}}}^\star,\text{M}^{\text{\student}})$ to synthesize a large set of codes as outputs along with probabilities. In our implementation, we further normalize these probabilities appropriately by considering the number of production rules involved.  In our experiments, we  sample a set of $1000$ codes and keep the codes with highest probabilities; Figures~\ref{fig:experiments.hoc4.generative.pcfg}~and~\ref{fig:experiments.hoc18.generative.pcfg} illustrate the Top-$3$ synthesized codes for two scenarios, obtained with this procedure. The Top-$1$ code is then used for generative performance evaluation. For the discriminative performance evaluation, we are already given a set of option codes; here we directly compute the likelihood of the provided options and then select one with the highest probability.

\section{Experimental Evaluation}\label{sec:experiments}
In this section, we expand on the evaluation presented in Section~\ref{sec:benchmark} and include results for \algoContinual{} and \algoPCFG{}. 

\subsection{Generative Performance}\label{sec:experiments.generative}
\textbf{Evaluation procedure.} As discussed in Section~\ref{sec:benchmark.measures}, we evaluate the generative performance of a technique in the following steps: (a) a scenario $(\task^{\text{ref}}, \code^{\text{\student}}_{\task^{\text{ref}}}, \task^{\text{tar}}, \text{\qmark})$ is picked; (b) the technique synthesizes \student's attempt, i.e., a program that \student{} would write if the system would assign $\task^{\text{tar}}$ to the student; (c) the generated code is scored on the $4$-point \textit{Likert scale}. The scoring step requires human-in-the-loop evaluation and involved an expert (different from the three experts that are part of \algoTutor{}). Overall, each technique is evaluated for $36$ unique scenarios in \benchmark{}---we selected $18$ scenarios per reference task by first picking one of the $3$ target tasks and then picking a student from one of the $6$ different types of behavior.
The final performance results in Table~\ref{table:benchmark.finalresults} are reported as an average across these scenarios; for \algoTutor{}, each of the three experts independently responded to these $36$ scenarios and the final performance is computed as a macro-average across experts.

\begin{figure*}[t!]
\centering
	\begin{subfigure}[b]{.25\textwidth}
	\centering
	{
			\begin{boxcode}{3.2cm}{0.80}{0.58}
				\\
				\\
				\\
				\vspace{2.6mm}			
				\hspace{1.1cm} {\fontsize{40}{1}\selectfont ?}
				\\
				\\
                \\
				\vspace{2.4mm}
			\end{boxcode}
			\vspace{-3.05mm}
			\caption{Attempt $\code^{\textnormal{\student}}_{\task^{\textnormal{4x}}}$}
			\label{fig:experiments.hoc4.generative.unknown}
	}
	\end{subfigure}
   \
	\begin{subfigure}[b]{.23\textwidth}
	\centering
	{
			\begin{boxcode}{3.4cm}{0.75}{0.80}
				\textcode{def }\DSLRun\textcode{()}\textcode{\{}\\
                \quad\DSLTurnRight\\
                \quad\DSLMove\\			
                \quad\DSLTurnLeft\\
                \quad\DSLMove\\
                \quad\DSLMove\\
                \quad\DSLMove\\
                \quad\DSLTurnLeft\\
                \quad\DSLMove\\
                \textcode{\}}
    		\end{boxcode}
			\vspace{-3.05mm}
			\caption{Solution $\code^{\star}_{\task^{\textnormal{4x}}}$}		
			\label{fig:experiments.hoc4.generative.solution}
	}
	\end{subfigure}
   \   
   \begin{subfigure}[b]{.23\textwidth}
	\centering
	{
		\begin{boxcode}{3.4cm}{0.75}{0.80}
			\textcode{def }\DSLRun\textcode{()}\textcode{\{}\\
            \quad\DSLTurnRight\\
            \quad\DSLMove\\			
            \quad\DSLTurnLeft\\
            \quad\DSLMove\\
            \quad\DSLMove\\
            \quad\DSLMove\\
            \quad\DSLMove\\
            \textcode{\}}
            \vspace{3.6mm}
		\end{boxcode}
		\vspace{-3.05mm}
		\caption{Benchmark code}
		\label{fig:experiments.hoc4.generative.benchmark}

	}
	\end{subfigure}	   
   \ 
   \begin{subfigure}[b]{.23\textwidth}
	\centering
	{
    		\begin{boxcode}{3.4cm}{0.75}{0.80}
    		\textcode{def }\DSLRun\textcode{()}\textcode{\{}\\
                \quad\DSLTurnRight\\
           	    \quad\DSLMove\\
                \quad\DSLMove\\
           	    \quad\DSLMove\\
           	    \quad\DSLMove\\
           	    \quad\DSLTurnLeft\\
           	    \quad\DSLMove\\
            \textcode{\}}
            \vspace{4mm}
    		\end{boxcode}						
			\vspace{-3.05mm}
    		\caption{\algoTutor} 
    		\label{fig:experiments.hoc4.generative.tutor}
	}
	\end{subfigure}
   \
		\begin{subfigure}[b]{0.49\textwidth}
		\vspace{2mm}
		{
			\begin{minipage}{0.31\textwidth}	
			\centering
			{
			\begin{boxcode}{3.4cm}{0.75}{0.80}
				\textcode{def }\DSLRun\textcode{()}\textcode{\{}\\
                \quad\DSLTurnRight\\
                \quad\DSLMove\\
                \quad\DSLMove\\
                \quad\DSLTurnLeft\\
	            \quad\DSLMove\\
	            \quad\DSLMove\\
                \textcode{\}}\\
                \\
			\end{boxcode}
    		}
    		\end{minipage}
    		\
			\begin{minipage}{0.31\textwidth}	
			\centering
			{
				\begin{boxcode}{3.4cm}{0.75}{0.80}
        		\textcode{def }\DSLRun\textcode{()}\textcode{\{}\\

                \quad\DSLTurnRight\\
        		\quad\DSLMove\\			
                \quad\DSLMove\\			
                \quad\DSLMove\\
                \quad\DSLTurnLeft\\
                \quad\DSLMove\\
                \textcode{\}}\\
                \\
				\end{boxcode}				
    		}
    		\end{minipage}
    		\
			\begin{minipage}{0.31\textwidth}	
			\centering
			{
				\begin{boxcode}{3.4cm}{0.75}{0.80}
                \textcode{def }\DSLRun\textcode{()}\textcode{\{}\\
                \quad\DSLTurnRight\\
        	    \quad\DSLMove\\
        	    \quad\DSLTurnLeft\\
        		\quad\DSLMove\\			
                \quad\DSLMove\\
                \quad\DSLMove\\
                \textcode{\}}
                \\
                \\
				\end{boxcode}
    		}
    		\end{minipage}
    		\
			\caption{\algoContinual{} -- Top-$3$ synthesized codes in decreasing likelihood}
			\label{fig:experiments.hoc4.generative.continual}
		}
		\end{subfigure}
	    \  \
		\begin{subfigure}[b]{0.49\textwidth}
		\vspace{2mm}
		{
			\begin{minipage}{0.31\textwidth}	
			\centering
			{
			\begin{boxcode}{3.4cm}{0.75}{0.80}
				\textcode{def }\DSLRun\textcode{()}\textcode{\{}\\
                \quad\DSLMove\\			
                \quad\DSLMove\\
                \quad\DSLMove\\
                \quad\DSLMove\\
	            \quad\DSLMove\\
                \textcode{\}}
                \\
                \\
				\vspace{3mm}
			\end{boxcode}
    		}
    		\end{minipage}
    		\ 
			\begin{minipage}{0.31\textwidth}	
			\centering
			{
				\begin{boxcode}{3.4cm}{0.75}{0.80}
				\textcode{def }\DSLRun\textcode{()}\textcode{\{}\\
                \quad\DSLMove\\			
                \quad\DSLMove\\
                \quad\DSLMove\\
                \quad\DSLMove\\
	            \quad\DSLTurnLeft\\
	            \quad\DSLMove\\
                \textcode{\}}
                \\
                \\
				\end{boxcode}				
    		}
    		\end{minipage}
    		\ 
			\begin{minipage}{0.31\textwidth}	
			\centering
			{
				\begin{boxcode}{3.4cm}{0.75}{0.80}
				\textcode{def }\DSLRun\textcode{()}\textcode{\{}\\
	            \quad\DSLTurnRight\\
                \quad\DSLMove\\			
                \quad\DSLMove\\
                \quad\DSLMove\\
                \quad\DSLMove\\
	            \quad\DSLMove\\
                \textcode{\}}
                \\
                \\
				\end{boxcode}
    		}
    		\end{minipage}
    		\
			\caption{\algoPCFG{} -- Top-$3$ synthesized codes in decreasing likelihood}
			\label{fig:experiments.hoc4.generative.pcfg}
		}
		\end{subfigure}	    
	\caption{Illustration of the qualitative results in terms of the generative objective for the scenario in Figure~\ref{fig:intro.hoc4}. \textbf{(a)} The goal is to synthesize the student \student{}'s behavior on the target task $\task^{\textnormal{4x}}$. \textbf{(b)} Solution code $\code^{\star}_{\task^{\textnormal{4x}}}$ for the target task. \textbf{(c)} Code provided in the benchmark as a possible answer for this scenario. \textbf{(d)} Code provided by one of the human experts. \textbf{(e, f)} Codes synthesized by our techniques \algoContinual{} and \algoPCFG{}---Top-$3$ synthesized codes in decreasing likelihood are provided here. See Section~\ref{sec:experiments.generative} for details.
	}
	\label{fig:experiments.hoc4.generative}
\end{figure*}



\begin{figure*}[t!]
\centering
	\begin{subfigure}[b]{.25\textwidth}
	\centering
	{
			\begin{boxcode}{3.15cm}{0.80}{0.58}
				\\
				\\
				\\
				\vspace{4.5mm}			
				\hspace{1.1cm} {\fontsize{40}{1}\selectfont ?}
				\\
				\\
				\\
				\vspace{4.6mm}
			\end{boxcode}
			\vspace{-3.05mm}
			\caption{Attempt $\code^{\textnormal{\student}}_{\task^{\textnormal{18x}}}$}	
			\label{fig:experiments.hoc18.generative.unknown}
	}
	\end{subfigure}
   \
   \begin{subfigure}[b]{.23\textwidth}
	\centering
	{
			\begin{boxcode}{3.4cm}{0.72}{0.57}
            \textcode{def }\DSLRun\textcode{()}\textcode{\{}\\
        	\quad\DSLMove\\
        	\quad\DSLMove\\
        	\quad\DSLTurnLeft\\
        	\quad\DSLRepeatUntil\textcode{(\DSLBoolGoal)}\textcode{\{}\\
       	 	\quad\quad\DSLIf\textcode{(\DSLBoolPathRight)}\textcode{\{}\\
        	\quad\quad\quad\DSLTurnRight\\
            \quad\quad\quad\DSLMove\\
        	\quad\quad\textcode{\}}\\
        	\quad\quad\DSLElse\textcode{\{}\\
        	\quad\quad\quad\DSLMove\\
       	 	\quad\quad\textcode{\}}\\
        	\quad\textcode{\}}\\
        	\textcode{\}}
            \vspace{-2mm}
			\end{boxcode}
			\vspace{-3.05mm}
			\caption{Solution $\code^{\star}_{\task^{\textnormal{18x}}}$} 
			\label{fig:experiments.hoc18.generative.solution}
	}
	\end{subfigure}	
   \
   \begin{subfigure}[b]{.23\textwidth}
	\centering
	{

			\begin{boxcode}{3.4cm}{0.72}{0.80}
           	    \textcode{def }\DSLRun\textcode{()}\textcode{\{}\\
                \quad\DSLRepeatUntil\textcode{(\DSLBoolGoal)}\textcode{\{}\\
                    \quad\quad\DSLMove\\
                    \quad\quad\DSLTurnLeft\\
               	    \quad\quad\DSLMove\\
                    \quad\quad\DSLTurnRight\\
               	    \quad\quad\DSLMove\\                    
                \quad\textcode{\}}\\
                \textcode{\}}%
                \\
                \\
                \\
				\vspace{-4mm}
			\end{boxcode}
			\vspace{-3.05mm}
			\caption{Benchmark code} 
			\label{fig:experiments.hoc18.generative.benchmark}
	}
	\end{subfigure}	   
   \ \ \ 
   \begin{subfigure}[b]{.23\textwidth}
	\centering
	{
				\begin{boxcode}{3.4cm}{0.75}{0.76}
				\textcode{def }\DSLRun\textcode{()}\textcode{\{}\\
                \quad\DSLRepeatUntil\textcode{(\DSLBoolGoal)}\textcode{\{}\\
               	    \quad\quad\DSLMove\\
                    \quad\quad\DSLMove\\
               	    \quad\quad\DSLTurnLeft\\
                    \quad\quad\DSLMove\\
               	    \quad\quad\DSLMove\\
               	    \quad\quad\DSLTurnRight\\
               	    \quad\quad\DSLMove\\
               	    \quad\quad\DSLMove\\
                \quad\textcode{\}}\\
                \textcode{\}}
				\end{boxcode}	
			\vspace{-3.05mm}
			\caption{\algoTutor{}} 
			\label{fig:experiments.hoc18.generative.tutor}
	}
	\end{subfigure}
   \ 
		\begin{subfigure}[b]{0.49\textwidth}
		\vspace{2mm}
		{
			\begin{minipage}{0.31\textwidth}	
			\centering
			{
			\begin{boxcode}{3.4cm}{0.75}{0.80}
				\textcode{def }\DSLRun\textcode{()}\textcode{\{}\\
                \quad\DSLRepeatUntil\textcode{(\DSLBoolGoal)}\textcode{\{}\\
               	    \quad\quad\DSLMove\\
                    \quad\quad\DSLTurnLeft\\
               	    \quad\quad\DSLMove\\
                    \quad\quad\DSLTurnLeft\\
               	    \quad\quad\DSLMove\\
                \quad\textcode{\}}\\
                \textcode{\}}
                \\
                \\
			\end{boxcode}
    		}
    		\end{minipage}
    		\
			\begin{minipage}{0.31\textwidth}	
			\centering
			{
				\begin{boxcode}{3.4cm}{0.75}{0.80}
				\textcode{def }\DSLRun\textcode{()}\textcode{\{}\\
                \quad\DSLRepeatUntil\textcode{(\DSLBoolGoal)}\textcode{\{}\\
               	    \quad\quad\DSLMove\\
               	    \quad\quad\DSLTurnLeft\\
                       \quad\quad\DSLMove\\
                \quad\textcode{\}}\\
                \textcode{\}}
                \\
                \\
                \\
                \\
				\end{boxcode}				
    		}
    		\end{minipage}
    		\
			\begin{minipage}{0.31\textwidth}	
			\centering
			{
				\begin{boxcode}{3.4cm}{0.75}{0.80}
				\textcode{def }\DSLRun\textcode{()}\textcode{\{}\\
                \quad\DSLRepeatUntil\textcode{(\DSLBoolGoal)}\textcode{\{}\\
               	    \quad\quad\DSLMove\\
                    \quad\quad\DSLTurnLeft\\
               	    \quad\quad\DSLMove\\
                    \quad\quad\DSLTurnLeft\\
                \quad\textcode{\}}\\
                \textcode{\}}
                \\
                \\
                \\

                \end{boxcode}
    		}
    		\end{minipage}
    		\
			\caption{\algoContinual{} -- Top-3 synthesized codes in decreasing likelihood}
			\label{fig:experiments.hoc18.generative.continual}
		}
		\end{subfigure}		
	\ \ 
	\begin{subfigure}[b]{0.49\textwidth}
		\vspace{2mm}
		{
			\begin{minipage}{0.31\textwidth}	
			\centering
			{
			\begin{boxcode}{3.4cm}{0.75}{0.80}
				\textcode{def }\DSLRun\textcode{()}\textcode{\{}\\
           	    \quad\DSLMove\\
                \quad\DSLMove\\                
                \quad\DSLTurnLeft\\
           \quad\DSLRepeatUntil\textcode{(\DSLBoolGoal)}\textcode{\{}\\
                    \quad\quad\DSLTurnRight\\
               	    \quad\quad\DSLMove\\
               	    \quad\quad\DSLMove\\
                \quad\textcode{\}}\\
                \textcode{\}}
           		\vspace{3mm}
			\end{boxcode}
    		}
    		\end{minipage}
    		\
			\begin{minipage}{0.31\textwidth}	
			\centering
			{
				\begin{boxcode}{3.4cm}{0.75}{0.80}
				\textcode{def }\DSLRun\textcode{()}\textcode{\{}\\
           	    \quad\DSLMove\\
                \quad\DSLMove\\                
                \quad\DSLTurnLeft\\
           \quad\DSLRepeatUntil\textcode{(\DSLBoolGoal)}\textcode{\{}\\
           	    \quad\quad\DSLMove\\
                    \quad\quad\DSLTurnRight\\
               	    \quad\quad\DSLMove\\
                     \quad\quad\DSLMove\\
                \quad\textcode{\}}\\
                \textcode{\}}
				\end{boxcode}				
    		}
    		\end{minipage}
    		\
			\begin{minipage}{0.31\textwidth}	
			\centering
			{
				\begin{boxcode}{3.4cm}{0.75}{0.80}
    			\textcode{def }\DSLRun\textcode{()}\textcode{\{}\\
           	    \quad\DSLMove\\
                \quad\DSLMove\\                
                \quad\DSLTurnLeft\\
           \quad\DSLRepeatUntil\textcode{(\DSLBoolGoal)}\textcode{\{}\\
                    \quad\quad\DSLTurnRight\\
               	    \quad\quad\DSLMove\\
               	    \quad\quad\DSLMove\\
                    \quad\quad\DSLMove\\
                \quad\textcode{\}}\\
                \textcode{\}}
			\end{boxcode}
    		}
    		\end{minipage}
    		\
			\caption{\algoPCFG{} -- Top-3 synthesized codes in decreasing likelihood}
			\label{fig:experiments.hoc18.generative.pcfg}
		}
		\end{subfigure}	
	\caption{Analogous to Figure~\ref{fig:experiments.hoc4.generative}, here we illustrate results in terms of the generative objective for the scenario in Figure~\ref{fig:intro.hoc18}.
	}
	\label{fig:experiments.hoc18.generative}
\end{figure*}


\textbf{Quantitative results.} Table~\ref{table:benchmark.finalresults} expands on Table~\ref{table:benchmark.initialresults} and reports results on the generative performance per reference task for different techniques. As noted in Section~\ref{sec:benchmark.initialresults}, the simple baselines (\algoRandom, \algoEditDist, \algoEmbDist) do not have a synthesis capability and hence have a score $1.00$. \algoTutor{}, i.e., human experts, achieves the highest performance with aggregated scores of $3.85$ and $3.91$ for two reference tasks respectively; as mentioned in Table~\ref{table:benchmark.initialresults}, these scores are reported as an average over scores achieved by three different experts. \algoPCFG{} also achieves high performance with aggregated scores of $3.72$ and $3.83$---only slightly lower than that of \algoTutor{} and these gaps are not statistically significant w.r.t. $\chi^2$ tests \cite{cochran1952chi2}. The high performance of \algoPCFG{} is expected given its knowledge about types of students in \benchmark{} and the expert domain knowledge inherent in its design. \algoContinual{} improves upon simple baselines and achieves aggregated scores of $3.28$ and $2.94$; however, this performance is significantly worse ($p\leq0.001$) compared to that of \algoPCFG{}  and \algoTutor{} w.r.t. $\chi^2$ tests.\footnote{$\chi^2$ tests reported here are conducted based on aggregated data across both the reference tasks.}
%

\textbf{Qualitative results.} Figures~\ref{fig:experiments.hoc4.generative}~and~\ref{fig:experiments.hoc18.generative} illustrate the qualitative results in terms of the generative objective for the scenarios in Figures~\ref{fig:intro.hoc4}~and~\ref{fig:intro.hoc18}, respectively. As can be seen in Figures~\ref{fig:experiments.hoc4.generative.tutor}~and~\ref{fig:experiments.hoc18.generative.tutor}, the codes generated by human experts in \algoTutor{} are high-scoring w.r.t. our $4$-point \textit{Likert scale}, and are slight variations of the \emph{ground-truth} codes in \benchmark{}  shown in Figures~\ref{fig:experiments.hoc4.generative.benchmark}~and~\ref{fig:experiments.hoc18.generative.benchmark}. 
Figures~\ref{fig:experiments.hoc4.generative.pcfg}~and~\ref{fig:experiments.hoc18.generative.pcfg} show the Top-$3$ codes synthesized by \algoPCFG{} for these two scenarios -- these codes are also  high-scoring w.r.t. our $4$-point \textit{Likert scale}. 
In contrast, for the scenario in Figure~\ref{fig:intro.hoc18}, the Top-$3$ codes synthesized by \algoContinual{} in Figure~\ref{fig:experiments.hoc18.generative.continual} only capture the elements of the student's behavior in $\code^{\text{\student}}_{\task^{\textnormal{ref}}}$ and miss the elements of the target task $\task^{\textnormal{tar}}$.

\subsection{Discriminative Performance}\label{sec.experiments.discriminative}
\label{sec:experiments:dicriminative}
\textbf{Evaluation procedure: Creating instances.} As discussed in Section~\ref{sec:benchmark.measures}, we evaluate the discriminative performance of a technique  in the following steps: (a) a discriminative instance is created with a scenario $(\task^{\text{ref}}, \code^{\text{\student}}_{\task^{\text{ref}}}, \task^{\text{tar}}, \text{\qmark})$  picked from the benchmark and $10$ code options created automatically; (b) the technique chooses one of the options as \student's attempt; (c) the chosen option is scored either $100.0$ when correct, or $0.0$ otherwise. We create a number of discriminative instances for evaluation, and then compute an average predictive accuracy in the range $[0.0, 100.0]$. We note that the number of discriminative instances can be much larger than the number of scenarios because of the variability in creating $10$ code options. When sampling large number of instances in our experiments, we ensure that all target tasks and behavior types are represented equally.  

\textbf{Evaluation procedure: Details about final performance.} For $\algoTutor{}$,  we perform evaluation on a small set of $72$ instances ($36$ instances per reference task), to reduce the effort for human experts. The final performance results for $\algoTutor{}$ in Table~\ref{table:benchmark.finalresults} are reported as an average predictive accuracy across the evaluated instances---each of the three experts independently responded to the instances and the final performance is computed as a macro-average across experts.
%
Next, we provide details on how the final performance results are computed for the techniques \algoRandom, \algoEditDist, \algoEmbDist, \algoContinual{}, and \algoPCFG{}. For these techniques, we perform $\numEval=5$ independent evaluation rounds, and report results as a macro-average across these rounds; these rounds are also used for statistical significance tests. Within one round, we create a set of $720$ instances ($360$ instances per reference task).  To allow hyperparameter tuning by techniques, we apply a cross-validation procedure on the $360$ instances per reference task by creating $10$-folds whereby $1$ fold is used to tune hyperparameters and $9$ folds are used to measure performance. Within a round, the performance results are computed as an average predictive accuracy across the  evaluated instances.

\textbf{Quantitative results.}
Table~\ref{table:benchmark.finalresults} reports results on the discriminative performance per reference task for different techniques. As noted in Section~\ref{sec:benchmark.initialresults}, the initial results showed a huge gap between the human experts (\algoTutor{}) and simple baselines (\algoRandom, \algoEditDist, \algoEmbDist). As can be seen in Table~\ref{table:benchmark.finalresults}, our proposed  techniques (\algoContinual{} and \algoPCFG{}) have  reduced this performance gap w.r.t. \algoTutor{}. \algoPCFG{} achieves high performance compared to simple baselines and \algoContinual{}; moreover, on the reference task $\task^\text{4}$, its performance ($87.17$) is close to that of \algoTutor{} ($89.81$).  The high performance of \algoPCFG{} is partly due to its access to types of students in \benchmark{}; in fact, this information is used only by \algoPCFG{} and is not even available to human experts in \algoTutor{}---see column ``Student types'' in Table~\ref{table:benchmark.finalresults}. \algoContinual{} outperformed simple baselines on the reference task $\task^\text{18}$; however, its performance is below \algoPCFG{} and \algoTutor{} for both the reference tasks. For the three techniques \algoContinual{}, \algoPCFG{}, and \algoEmbDist,  we did  statistical significance tests based on results from $\numEval=5$ independent rounds w.r.t. Tukey's HSD test \cite{Tukey1949ComparingIM}, and obtained the following: (a) the performance of \algoContinual{} is significantly better than \algoEmbDist{} on the reference task $\task^\text{18}$ ($p \leq 0.001$); (b) the performance of \algoPCFG{} is significantly better than \algoContinual{} and \algoEmbDist{} on both the reference tasks ($p \leq 0.001$).

%

\section{Conclusions and Outlook}\label{sec:conclusions}
We investigated student modeling in the context of block-based visual programming environments, focusing on the ability to automatically infer students' misconceptions and synthesize their expected behavior. We introduced a novel benchmark, \benchmark{}, to objectively measure the generative as well as the discriminative performance of different techniques. The gap in performance between human experts (\algoTutor) and our techniques (\algoContinual{}, \algoPCFG) highlights the challenges in synthesizing student attempts for programming tasks. We believe that the benchmark will facilitate further research in this crucial area of student modeling for block-based visual programming environments.

There are several important directions for future work, including but not limited to: (a) incorporating more diverse tasks and student misconceptions in the benchmark; (b) scaling up the benchmark and creating a competition with a public leaderboard to facilitate research; (c) developing new neuro-symbolic synthesis techniques that can get close to the performance of \algoTutor{} without relying on expert inputs; (d) applying our methodology to other programming environments (e.g., Python programming).

\section{Acknowledgments}
Funded/Co-funded by the European Union (ERC, TOPS, 101039090). Views and opinions expressed are however those of the author(s) only and do not necessarily reflect those of the European Union or the European Research Council. Neither the European Union nor the granting authority can be held responsible for them.

\bibliographystyle{abbrv}
\bibliography{main}

\end{document}